\documentclass[sigconf]{acmart}
\usepackage{booktabs} 
\usepackage[ruled]{algorithm2e} 
\usepackage{algpseudocode} 
\usepackage{graphicx}
\usepackage{subcaption}
\usepackage{epsfig}   
\usepackage{xcolor}
\usepackage{amsthm}
\usepackage{multirow}
\usepackage{amssymb}
\usepackage{appendix}
\usepackage{xcolor}
\usepackage{pdfpages}
\usepackage{url}
\usepackage{balance}
\include{pythonlisting}
\usepackage{pifont}
\newcommand{\sect}{\textsection}

\newcommand{\chris}[1]{{\color{black}{#1}}}

\newcommand{\sysname}{\texttt{AutoTune}\xspace}
\newcommand{\eat}[1]{}

\theoremstyle{definition}

\SetAlFnt{\small}
\SetAlCapFnt{\small}
\SetAlCapNameFnt{\small}
\SetAlCapHSkip{0pt}
\IncMargin{-\parindent}
\SetKwInput{kwInit}{Initialize}
\acmJournal{IMWUT}
\acmVolume{0}
\acmNumber{0}
\acmArticle{0}
\acmYear{2018}
\acmMonth{0}
\acmArticleSeq{0}
\acmPrice{15.00}

\copyrightyear{2019}
\acmYear{2019} 
\setcopyright{iw3c2w3}
\acmConference[WWW '19]{Proceedings of the 2019 World Wide Web Conference}{May 13--17, 2019}{San Francisco, CA, USA}
\acmBooktitle{Proceedings of the 2019 World Wide Web Conference (WWW'19), May 13--17, 2019, San Francisco, CA, USA}
\acmPrice{}
\acmDOI{10.1145/3308558.3313398}
\acmISBN{978-1-4503-6674-8/19/05}

\begin{document}

\title[AutoTune]{Autonomous Learning for Face Recognition in the Wild \\via Ambient Wireless Cues}
\author{Chris Xiaoxuan Lu}
\orcid{0000-0002-3733-4480}
\authornote{Both authors contributed equally to the paper.}
\affiliation{%
  \institution{University of Oxford}
}
\email{xiaoxuan.lu@cs.ox.ac.uk}

\author{Xuan Kan}
\authornotemark[1]
\affiliation{%
  \institution{Tongji University}
}
\email{kanxuan1996@gmail.com}

\author{Bowen Du}
\affiliation{%
  \institution{University of Warwick}
}
\email{B.Du@warwick.ac.uk}

\author{Changhao Chen}
\affiliation{%
  \institution{University of Oxford}
}
\email{changhao.chen@cs.ox.ac.uk}

\author{Hongkai Wen}
\affiliation{%
  \institution{University of Warwick}
}
\email{hongkai.wen@dcs.warwick.ac.uk}

\author{Andrew Markham}
\affiliation{%
  \institution{University of Oxford}
}
\email{andrew.markham@cs.ox.ac.uk}

\author{Niki Trigoni}
\affiliation{%
  \institution{University of Oxford}
}
\email{niki.trigoni@cs.ox.ac.uk}

\author{John A. Stankovic}
\affiliation{%
  \institution{University of Virginia}
}
\email{jas9f@virginia.edu}

\begin{abstract}
Facial recognition is a key enabling component for emerging Internet of Things (IoT) services such as smart homes or responsive offices. Through the use of deep neural networks, facial recognition has achieved excellent performance. However, this is only possibly when trained with hundreds of images of each user in different viewing and lighting conditions. Clearly, this level of effort in enrolment and labelling is impossible for wide-spread deployment and adoption. Inspired by the fact that most people carry smart wireless devices with them, e.g. smartphones, we propose to use this wireless identifier as a supervisory label. This allows us to curate a dataset of facial images that are unique to a certain domain e.g. a set of people in a particular office. This custom corpus can then be used to finetune existing pre-trained models e.g. FaceNet. However, due to the vagaries of wireless propagation in buildings, the supervisory labels are noisy and weak. We propose a novel technique, \sysname, which learns and refines the association between a face and wireless identifier over time, by increasing the inter-cluster separation and minimizing the intra-cluster distance. Through extensive experiments with multiple users on two sites, we demonstrate the ability of \sysname to design an environment-specific, continually evolving facial recognition system with entirely no user effort. 
\end{abstract}  



\thanks{}

\maketitle
%
\renewcommand{\shortauthors}{C. X. Lu et al.}
\section{Introduction} 
\label{sec:introduction}

Facial recognition and verification are key components of smart spaces, e.g., offices and buildings for determining who is where. Knowing this information allows a building management system to tailor ambient conditions to particular users, perform automated security (e.g., opening doors for the correct users without the need for a swipe card), and customize smart services (e.g., coffee dispensing). A vast amount of research over the past decades has gone into designing tailored systems for facial recognition and with the advent of deep learning, progress has accelerated. As an example of a state-of-the-art face recognizer, FaceNet achieves extremely high accuracies (e.g., 99.5\%) on very challenging datasets through the use of a low dimensional embedding, allowing similar faces to be clustered through their Euclidean distance \cite{schroff2015facenet,liu2017sphereface}. However, when directed transferred to operate in `in the wild', subject to variable lighting conditions, viewing angle and appearance changes, performance of off-the-shelf pre-trained classifiers degrades significantly, with accuracies around 15\% not being uncommon. The solution to this is to obtain a large, labelled corpus of data for a particular environment, with hundreds of annotated images per user. Given access to such a hypothetical dataset, it is then possible to fine-tune the pre-trained classifier on out-domain data to adapt to the new environment and achieve excellent performance.

However, the cost of labelling and updating the corpus (e.g. to enrol new users) is prohibitive for most critical applications and therefore, will naturally limit the use and uptake of facial recognition as a ubiquitous technology in emerging Internet of Things (IoT) applications. On the other side, people often, but not always, carry smart devices (phones, fitness devices etc). Wang et al. \cite{wang2018enabling} advocated that although these devices do not provide fine-grained enough positioning capability to act as a proxy for presence, they can be used to indicate that a user \emph{might} be present in an area with a co-located camera. In this work, we take a step forward and further utilize device presence as weak supervision signals for the purposes of fine-tuning a classifier. The goal now becomes how to take an arbitrary, pre-trained recognition network and tune it from a generic classifier to a highly specific classifier, optimized for a certain environment and group of people. We note that the aim is to make the network better and better at this specific goal, but it would likely perform poorly if transferred directly to a different environment. This is the antithesis of the conventional view of generalized machine learning, but is ideally suited for the problem of environment specific facial recognition, as opposed to generic facial recognition.

The technical challenge is that there is not a 1:1 mapping between a face and a wireless identity, rather we need to solve the association between a set of faces and a set of identities over many sessions or occasions. To further complicate the problem, the sets are not pure i.e. the set of faces can contain additional faces from people not of interest (e.g. visitors). Equally well, due to the vagaries of wireless transmission, the set of wireless identifiers will contain additional identifiers e.g. from people in the next office. Furthermore, it is also possible to have missing observations e.g. because a person was not facing the camera or because someone left their phone at home. 

In this work, we present {\sysname}, a system which can be used to gradually improve the performance of facial recognition systems in the wild, with zero user effort, tailoring them to the visual specifics of a particular smart space. We demonstrate state-of-the-art performance in real-world facial recognition through a number of experiments and trials.

In particular, our contributions are:
\begin{itemize}
\item We observe and prove that wireless signals of users' devices provide valuable, albeit noisy, clues for face recognition. Namely, wireless signals can serve as a weak label. Such weak labels can replace the human annotated face images in the wild to save intensive effort. 

\item We create {\sysname}, a novel pipeline to simultaneously label face images in the wild and adapt the pre-trained deep neural network to recognize the faces of users in new environments. The key idea is to repeat the face-identity association and network update in tandem. To cope with observation noise, we propose a novel probabilistic framework in {\sysname} and design a new stochastic center loss to enhance the robustness of network fine-tuning. 

\item We deployed {\sysname} in two real-world environments and experimental results demonstrate that {\sysname} is able to achieve $>0.85$ $F_1$ score of image labeling in both environments, outperforming the best competing approach by $>25\%$. Compared to the best competing approach, using the features extracted from the fine-tuned model and training a classifier based on the cross-modality labeled images can give a $\sim 19\%$ performance gain for online face recognition.

\end{itemize}

The rest of this paper is organized as follows. \sect\ref{sec:related_work} introduces the background of this work. System overview is given in \sect\ref{sec:overview}. We describe the {\sysname} solution in \sect\ref{sec:cross_modality_labeling} and \sect\ref{sec:model_update}. System implementation details are given in \sect\ref{sec:implementation}. The proposed approach is evaluated and compared with state of the art methods in \sect\ref{sec:evaluation}. Finally, we discuss and outlook future work in \sect\ref{sec:discussion_and_future_work} and conclude in \sect\ref{sec:conclusion}.
\section{Related Work} 
\label{sec:related_work}
\noindent \textbf{Deep face recognition: }
Face recognition is arguably one of the most active research areas in the past few years, with a vast corpus of face verification and recognition work \cite{parkhi2015deep,zhao2003face,tan2006face}. With the advent of deep learning, progress has accelerated significantly. Here we briefly overview state-of-the art work in Deep Face Recognition (DFR). Taigman et al. pioneered this research area and proposed DeepFace~\cite{taigman2014deepface}. It uses CNNs supervised by softmax loss, which essentially solves a multi-class classification problem. When introduced, DeepFace achieved the best performance on the Labeled Face in the Wild (LFW)~\cite{huang2007labeled} benchmark. Since then, many DFR systems have been proposed. In a series of papers~\cite{sun2014deep1,sun2014deep2} Sun et al. extended on DeepFace incrementally and steadily increased the recognition performance. A critical point in DFR happened in 2015, when researchers from Google~\cite{schroff2015facenet} used a massive dataset of 200 million face identities and 800 million image face pairs to train a CNN called Facenet, which largely outperformed prior art on the LFW benchmark when introduced. A point of difference is in their use of a ``triplet-based'' loss~\cite{cheng2016person}, that guides the network to learn both inter-class dispersion and inner-class compactness. Recently proposed RTFace \cite{wang2018enabling} not only achieves high recognition accuracy but also operates at the full frame rates of videos. Although the above methods have proven remarkably effective in face recognition, the training needs a vast amount of labeled images to train the supervised DFR network. A large amount of labeled data is not always achievable in a particular domain, and using a small amount of training data will incur poor generalization ability in the wild. 

\noindent \textbf{Cross-modality Matching: }
Cross-modal matching has received considerable attention in different research areas. Methods have been developed to establish mappings from images \cite{frome2013devise,karpathy2015deep,vinyals2015show} and videos \cite{venugopalan2014translating} to textual descriptions (e.g., captioning), developing image representation from sounds \cite{owens2016ambient,nagrani2018seeing}, recognizing speaker identities from Google calendar information \cite{lu2017scan}, and generating visual models from text \cite{zitnick2013learning}. In cross-modality matching between images and radio signals, however, work is very limited and all dedicated to trajectory tracking of humans \cite{teng2014ev,alahi2015rgb,papaioannou2015accurate}. The field of face recognition via wireless signals is an unexplored area.

\section{{\sysname} Overview} 
\label{sec:overview}

\begin{figure*}[ht]
	\begin{minipage}[b]{0.49\linewidth}
		\centering
		\includegraphics[width=\textwidth]{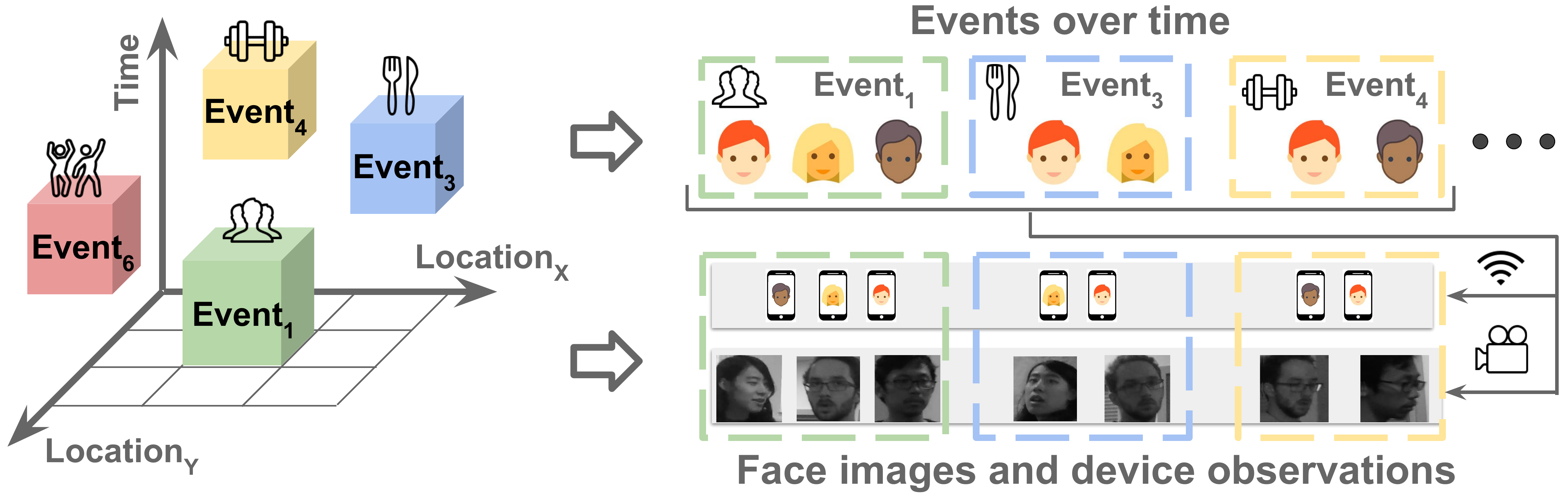}
		\caption{An illustrative example of \emph{events} and showing their linked \emph{observations}. An \emph{event} is uniquely determined by spatial-temporal attributes and its participants. An event is also linked with two types of observations. Face observation are images cropped from surveillance videos and device observation are sniffed MAC address of participants' devices.}
		\label{fig:event_def}
	\end{minipage}
	\hfill
	\begin{minipage}[b]{0.49\linewidth}
		\centering
		\includegraphics[width=\textwidth]{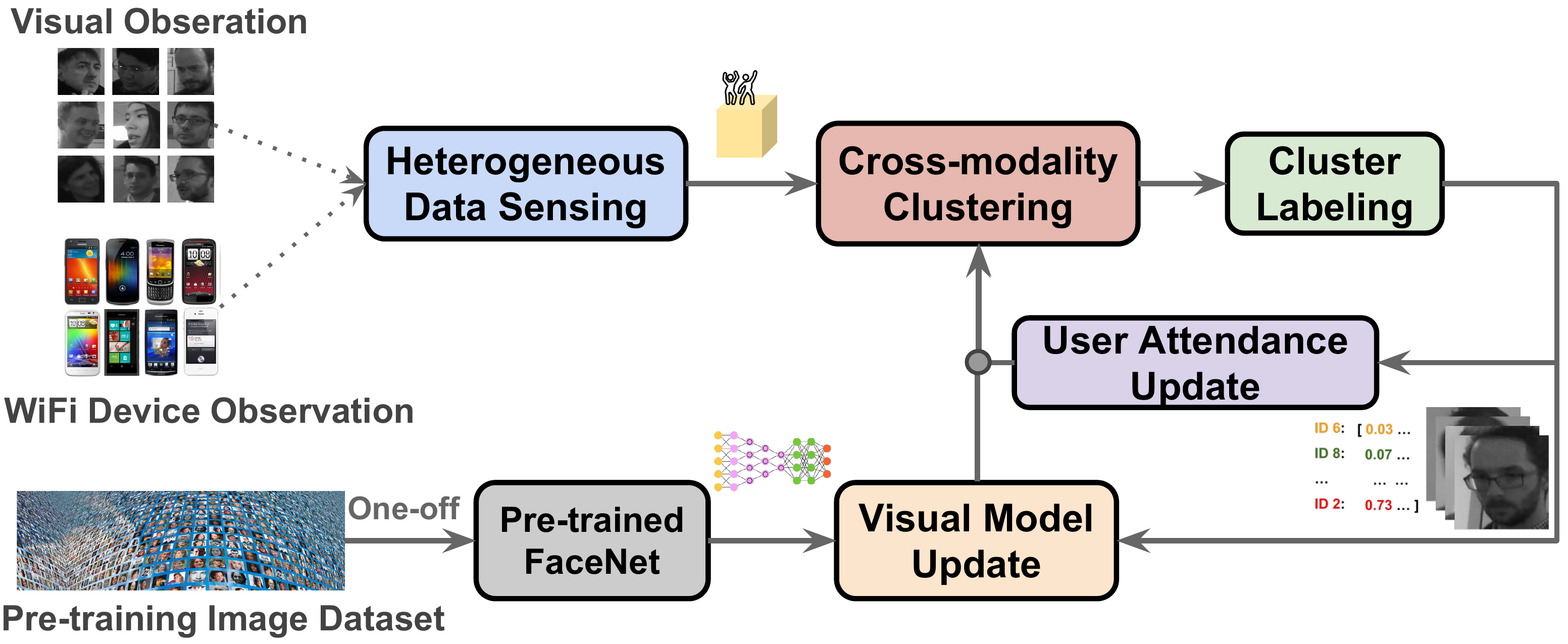}
		\caption{Workflow of {\sysname}. {\sysname} consists of 5 steps i) Heterogeneous Data Sensing ii) Cross-Modality Clustering iii) Cluster Labeling iv) Visual Model Update v) User Attendance Update. {\sysname} sequentially repeats the above five steps until the changes in the user attendance are negligible.}
		\label{fig:architecture}
	\end{minipage}
\end{figure*}

\subsection{System Model} 
\label{sub:system_model}
We consider a face recognition problem with $m$ people of interest (POI) and each subject owns one WiFi-enabled device, e.g., a smartphone. We denote the identity set as $\mathcal{Y} = \{y_j|j=1,2, \ldots, m\}$. The set of observed POI's devices in a particular environment is denoted by $\mathcal{L} = \{l_j|j=1,2, \ldots, m\}$, e.g., a set of MAC addresses. We assume the mapping from device MAC addresses $\mathcal{L}$ to the user identity $\mathcal{I}$ is known. A collection of face images $\mathcal{X} = \{x_j|j=1,2, \ldots, n\}$ is cropped from the surveillance videos in the same environment. Note that, to mimic the real-world complexity, our collection includes the faces of both POI and some non-POI, e.g., subjects with unknown device MAC addresses. We then assign face and device observations to different subsets based on their belonging events $\mathcal{E} = \{e_j|j=1,2, \ldots, h\}$. An event $e_j$ is the setting in which people interact with each other in a specific part of the environment for a given time interval. It is uniquely identified by three attributes: effective timeslot, location, and participants. Fig.~\ref{fig:event_def} demonstrates a few examples of events. Lastly, we also have a deep face representation model $f_{\theta}$ pre-trained on public datasets that contains no POI. Such model is trained with metric losses, e.g., triplet loss so that the learned features could bear a good property for clustering \cite{schroff2015facenet}.

In this sense, the problem addressed by {\sysname} is assigning IDs to images from noisy observations of images and WiFi MAC addresses, and using such learned ID-image associations to tune the pre-trained deep face representation model automatically.

\subsection{System Architecture} 
\label{sub:system_architecture}
\sysname is based on two key observations: i) although collected by different modalities, both face images and device MAC addresses are linked with the identities of users who attend certain events; and ii) the tasks of model tuning and face-identity association should not be dealt with separately, but rather progress in tandem. Based on the above insights, {\sysname} works as follows (see Fig~\ref{fig:architecture}):
\begin{itemize}
	\item \emph{Heterogeneous Data Sensing}. This module collects facial and WiFi data (device attendance observations) through surveillance cameras and WiFi sniffers\footnote{\url{https://www.wireshark.org/}} in a target environment. Given the face images and sniffed WiFi MAC addresses, {\sysname} first segments them into events based on the time and location they were captured.

	\item \emph{Cross-modality Labeling}. This module first clusters face images based on their appearance similarity computed by the face representation model, and also taking into account information on device attendance in events. Each image cluster should broadly correspond to a user, and the cluster's images are drawn from a set of events. We then assign each cluster to the user whose device has been detected in as similar as possible set of events. 

	\item \emph{Model Updates}. Once images are labeled with user identity labels, this module then fine-tunes the pre-trained face representation model. {\sysname} further uses the cluster labels to update our belief on which device (MAC address) has participated in each event. 
\end{itemize}

The sensing module is one-off and we will detail its implementation in \sect\ref{sub:heterogeneous_data_sensing}. Labeling and model update modules are iteratively repeated until the changes in the user attendance model become negligible. The tuned model derived in the last iteration is regarded as the one best adapted to POI recognition in the new environment.

\section{Cross-modality Labeling} 
\label{sec:cross_modality_labeling}

In this section, we introduce the labeling module in {\sysname}. The challenge in this module is that collected facial images and sniffed WiFi data are temporally unaligned. For example, detecting a device WiFi address does not imply that the device owner will be captured by the camera at the exact instant and vice versa. Such mismatches in cross-modality data distinguish our problem from prior sensor fusion problems, where both multiple sensors are observing a temporal evolving system. In order to tackle the above challenge, we leverage the diverse attendance patterns in events and use a two-step procedure in isolated. Images $\mathcal{X}$ are firstly grouped together into clusters across all sessions, and we then associate clusters with device IDs (i.e., labels) $\mathcal{L}$ based on their similarity in terms of event attendance.

\subsection{Cross-modality Clustering} 
\label{sub:cross_modality_clustering}

\noindent \textbf{Heterogeneous Features}. 
Given a pre-trained face representation model $f_{\theta}$, an image $x_i$ can be translated to the feature vector $\mathbf{z}_i$. Unlike conventional clustering that merely depends on the face feature similarity, {\sysname} merges face images \textit{across events} into a cluster (potentially belonging to the same subject) by incorporating attendance information as well. Recall that device attendance already reveals the identities of subjects (in the form of MAC addresses) in a particular event, and the captured images in the same event may contain the faces of these subjects as well. Despite the noise in observations, the overlapped subjects in different events can be employed as a prior that guides the image clustering. For example, if there are no shared MAC addresses sniffed in two events, then it is very likely that the face images captured in these two events should lie in different clusters. Formally, for an event $e_i$, we denote a device attendance vector as $\mathbf{u}_{k} = (u_k^1, u_k^2, \ldots, u_k^m)$, where $u_k^j=1$ if device $l_j$ is detected in event $e_k$. In this way, we could construct a heterogeneous feature $\widetilde{\mathbf{z}}_i = [\mathbf{z}_i, \mathbf{u}_{k}]$ for an image $x_i$ collected in the event $e_k$. Note that, as all images captured in a same event have the same attendance vector, this device attendance part are essentially enforced on cross-event image comparison. 

\noindent \textbf{Face Image Similarity}.
Given an image $x_i$ captured in event $e_k$ (i.e., $\widetilde{\mathbf{z}}_i = [\mathbf{z}_i, \mathbf{u}_{k}]$) and an image $x_j$ captured in event $e_p$ (i.e., $\widetilde{\mathbf{z}}_j = [\mathbf{z}_j, \mathbf{u}_{p}]$), the likelihood that two cross-event face images belong to the same subject is conditioned on two factors: i) the similarity of their feature representation between $\mathbf{z}_{i}$ and $\mathbf{z}_{j}$; and ii) the overlap ratio between the attendance observations in their corresponding events $\mathbf{u}_{k}$ and $\mathbf{u}_{p}$ respectively. The resulted joint similarity is a log likelihood $log(\mathbf{Pr}(x_i = x_j))$ defined as follows:

\begin{equation*}
\begin{split}
	\mathbf{Pr}(x_i = x_j) 
	& \propto \frac{exp(\beta * |\mathbf{u}_{k} \otimes \mathbf{u}_{p}|}{exp(\beta * |\mathbf{u}_{k} \oplus \mathbf{u}_{p}|)} * exp(-D(\mathbf{z}_i, \mathbf{z}_j)) \\
	log(\mathbf{Pr}(x_i = x_j)) & \propto \beta * \underbrace{(|\mathbf{u}_{k} \otimes \mathbf{u}_{p}| - |\mathbf{u}_{k} \oplus \mathbf{u}_{p}|)}_\text{attendance similarity} - \underbrace{D(\mathbf{z}_i, \mathbf{z}_j)}_\text{feature distance}
	\label{equ:id-ass_sim}
\end{split}
\end{equation*}

Here $\otimes$ and $\oplus$ are element-wise AND and OR, and $|\cdot|$ here is the $L^1$-norm. $z$ is the features transformed by the face representation model and $D$ is a distance measure between face features. $\beta$, analogous to the regularization parameter in composite loss functions, is a hyper-parameter that controls the contributions of the attendance assistance and feature similarity. The above derivation is inspired by the Jaccard coefficients, with the difference lying in the log function. The rationale behind the term $|\mathbf{u}_{k} \oplus \mathbf{u}_{p}|$ is that the more different subjects attending events, the more uncertain that any two images drawn across these events will point to the same subject. In contrast, when the intersection $|\mathbf{u}_{k} \otimes \mathbf{u}_{p}|$ is significant enough, the chance that these two images point to the same subject will become higher. This joint similarity can also be explained from a Bayesian perspective. The attendance similarity of two events can serve as a prior that two cross-event images belong to the same subject and the feature similarity can be seen as the likelihood. Together they determine the posterior probability that two cross-event images fall into the same cluster. Based on the above joint similarity, images across events are grouped into clusters $\mathcal{C} = \{c_i|i =1,2, \ldots, g\}$. We will soon discuss how to determine the number of clusters $g$ based on the complete set of MAC addresses $\mathcal{L}$ in the next section.

\begin{figure}[t]\centering
\includegraphics[width=\columnwidth]{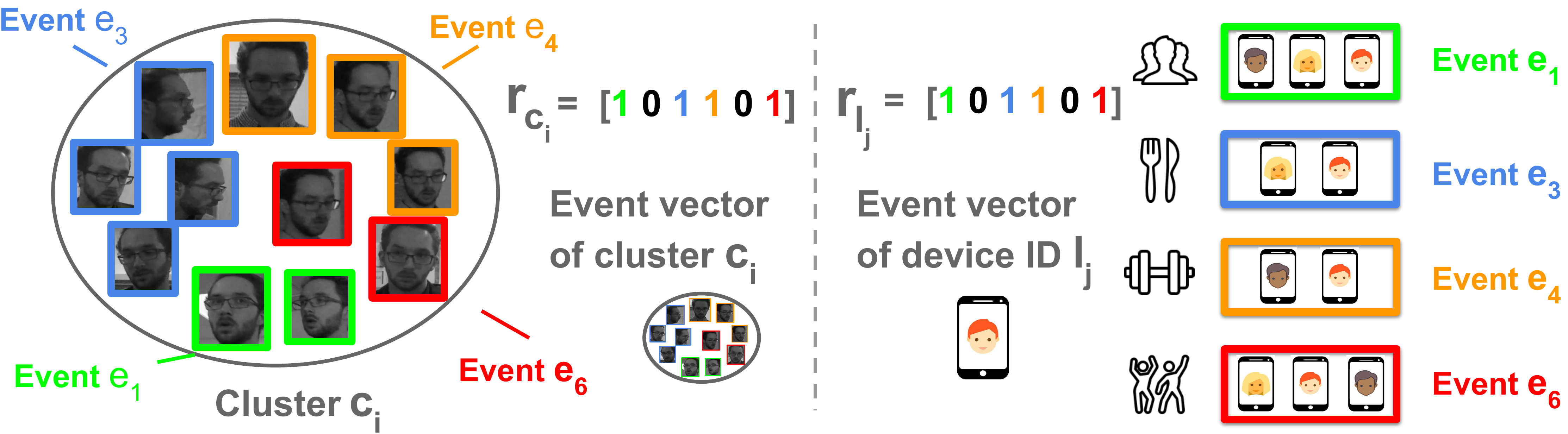}  
\caption{Event vectors of an image cluster $c_i$ and device ID $l_j$. The $k$th element of the vector $r_{c_i}^k$ is set to $1$ only if it contains images attached with the event $e_k$. The event vector $\mathbf{r}_{l_j}$ of the device ID $l_j$ is similarly developed. This insight leads to Eq.~\ref{eq:assoc}.} 
\label{fig:event_vector}
\end{figure}

\subsection{Cluster Labeling} 
\label{sub:cluster_labeling}

\noindent \textbf{Fine-grained ID Association}.
After clustering, an image cluster is linked with multiple events that are associated with its member images. Naturally, we introduce an event vector $\mathbf{r}_{c_i} = (r_{c_i}^1, r_{c_i}^2, \ldots, r_{c_i}^{h})$ for for an image cluster $c_i$, where $h$ is the total number of events. $r_{c_i}^k$ is set to 1 only if $c_i$ contains images from event $e_k$. Fig.~\ref{fig:event_vector} shows an example of how an event vector is developed. Similarly, for a device ID $l_j$, its corresponding event vector $\mathbf{r}_{l_j} = (r_{l_j}^1, r_{l_j}^2, \ldots, r_{l_j}^h)$ can be determined by inspecting its occurrences in all WiFi sniffing observations. $r_{l_j}^k$ is set to $1$ only if the device ID (MAC address) $l_j$ is detected in the event $e_k$. The intuition behind ID association is that a device and a face image cluster of the same subject should share the most consistent attendance pattern in events, reflected by the similarity of their event vectors. Based on this intuition, {\sysname} assigns clusters with device IDs based on the matching level of their event vectors. Formally, a matching problem of bipartite graph can be formulated as follows:
\begin{equation}\label{eq:assoc}
	\begin{split}
		& \mathcal{L}_A = \sum_{ij} a_{ij} (\mathbf{r}_{c_i} - \mathbf{r}_{l_j})^2 \\
		 & \sum_{1 \leq j \leq m}  a_{ij} = 1, \quad \forall i \in \{1, \ldots, g\}
	\end{split}
\end{equation}
where the solution of the binary variable $a_{ij}$ assigns a device ID to a cluster. We note that when $m<g$, {\sysname} adds dummy nodes to create the complete bipartite graph. Then the complete bipartite graph problem can be solved by the Hungarian algorithm \cite{jonker1986improving}.

\begin{figure*}[t]
	\centering
	\begin{subfigure}{0.25\textwidth}
	\includegraphics[width=0.9\linewidth]{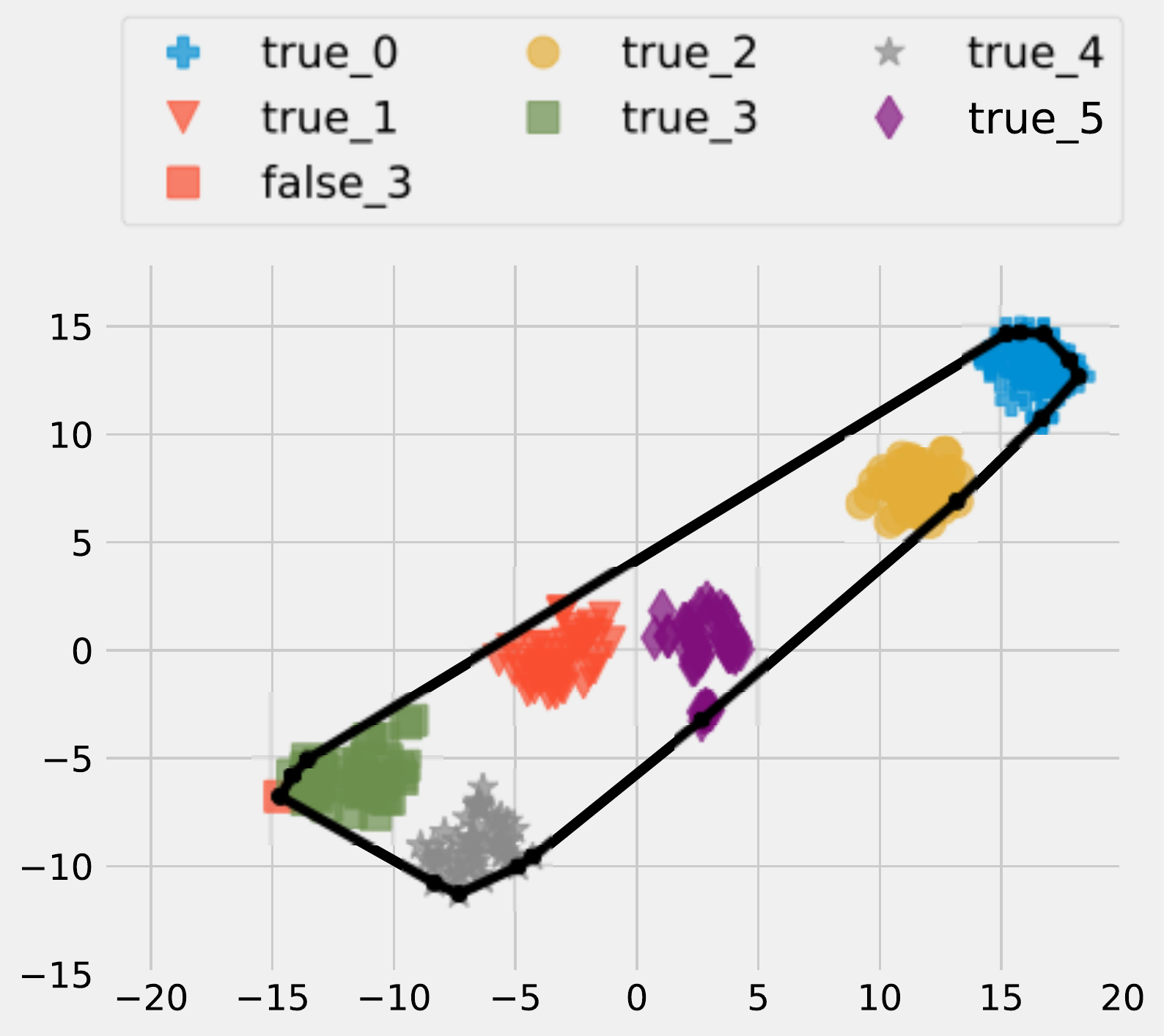}
	\caption{Pre-trained Model} \label{fig:tnse_pretrain}
	\end{subfigure}\hspace*{\fill}
	\begin{subfigure}{0.25\textwidth}
	\includegraphics[width=0.9\linewidth]{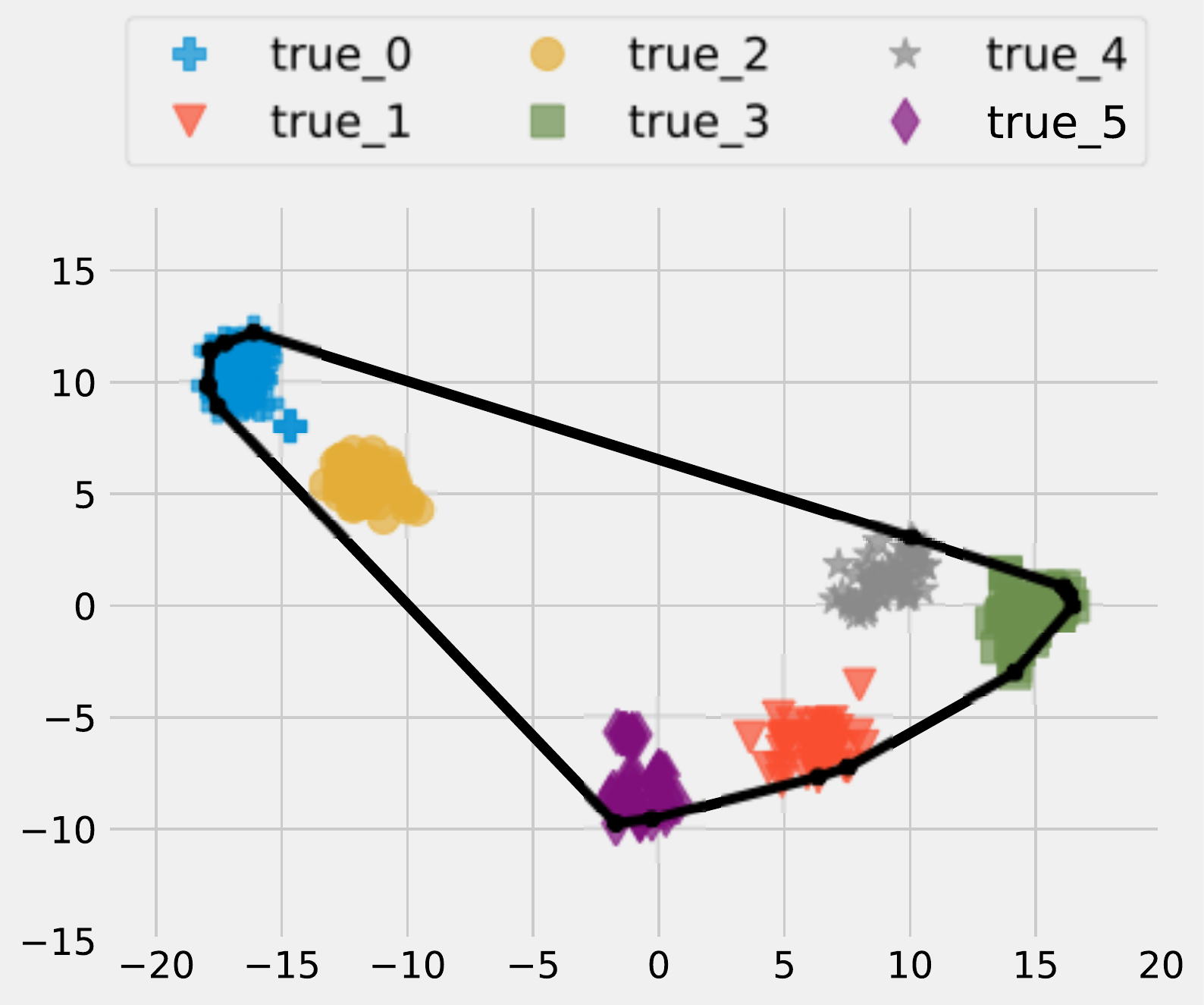}
	\caption{1st iteration}
	\end{subfigure}\hspace*{\fill}
	\centering
	\begin{subfigure}{0.25\textwidth}
	\includegraphics[width=0.9\linewidth]{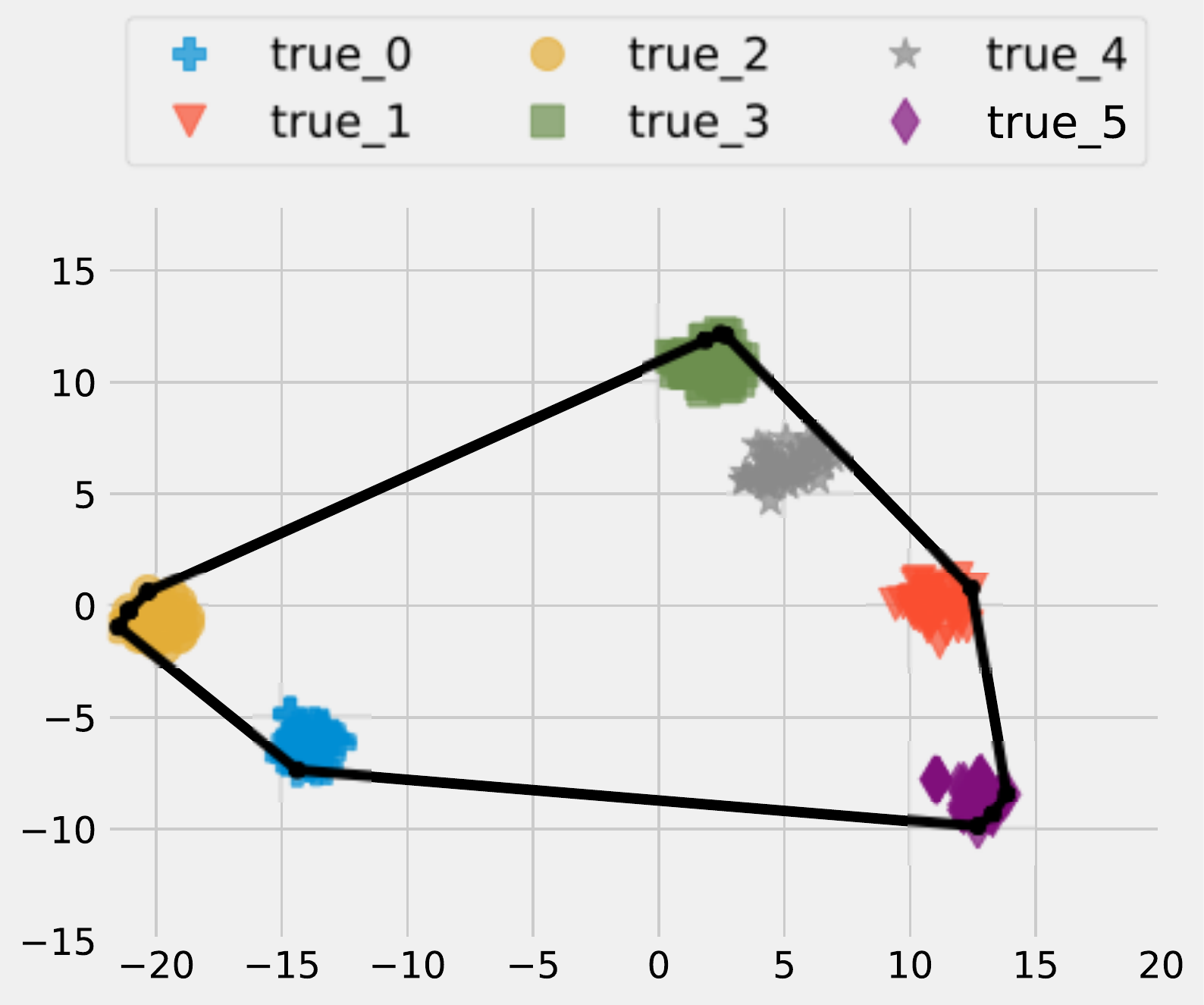}
	\caption{2nd iteration} 
	\end{subfigure}\hspace*{\fill}
	\begin{subfigure}{0.25\textwidth}
	\includegraphics[width=0.9\linewidth]{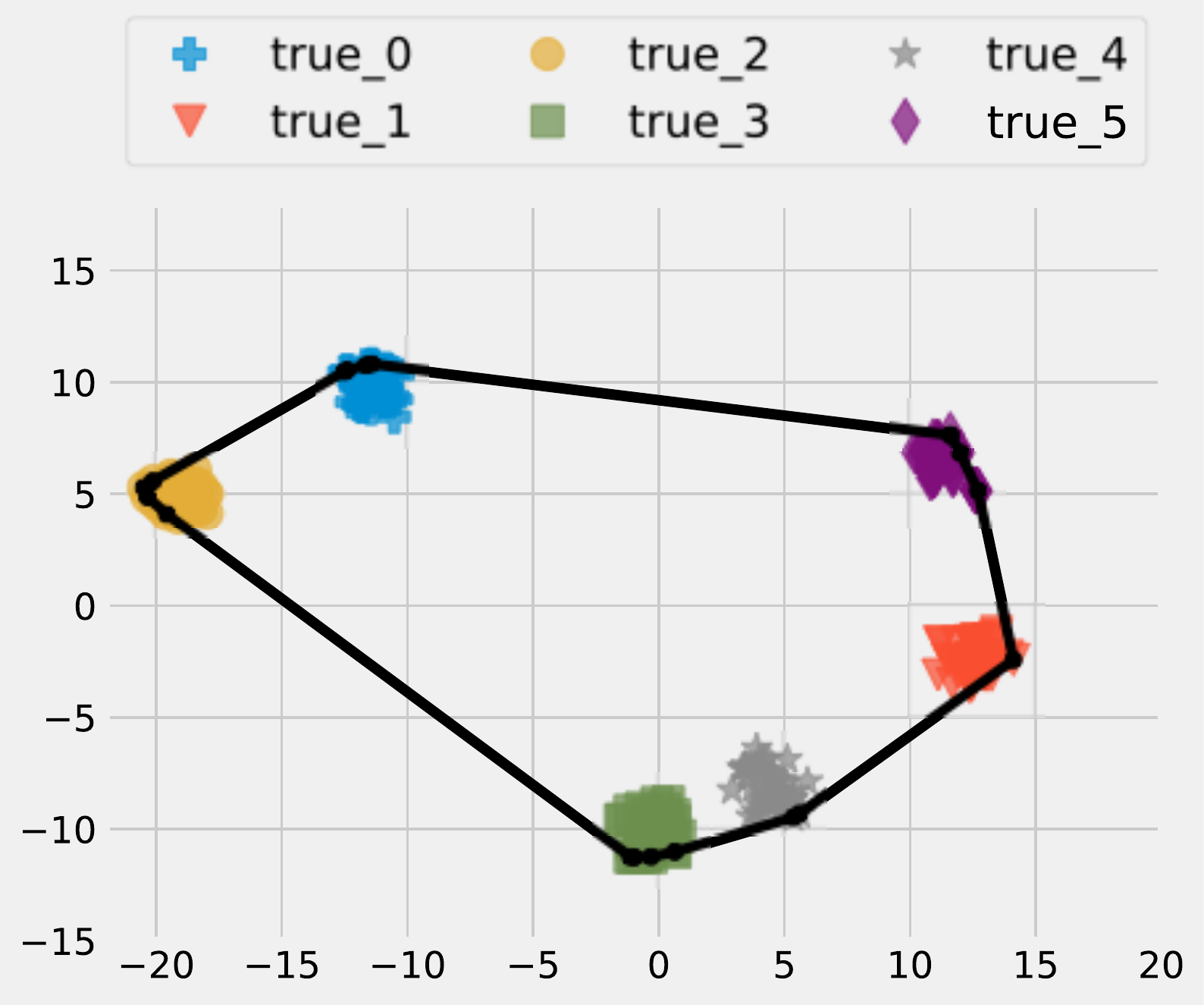}
	\caption{3rd iteration} 
	\end{subfigure}

\caption{An example of visual model update (\sect\ref{sub:visual_model_update}) in {\sysname} with a six-subject subset. Note that the pre-trained model in Fig.~\ref{fig:tnse_pretrain} has a narrow convex hull (black lines), clusters are dispersed and there is mis-classification (see false$\_3$ in Fig.~\ref{fig:tnse_pretrain}) as the pre-trained features are not discriminative enough to differentiate unseen subjects in a new environments. Over time, clusters become more compact and move further away from each other, leading to purer clustering results.} \label{fig:model_update}
\end{figure*}

\noindent \textbf{Probabilistic Labeling via Soft Voting}.
We now obtain the association between face images and device IDs. However, in practice, the number of clusters $g$ can vary due to the captured face images outside the POI, e.g., face images of short-term visitors. The choice of $g$ has a significant impact on the performance of clustering \cite{fraley1998many} which could further affect the following fine-tuning performance. To cope with this issue, {\sysname} sets the number of clusters $g$ greater than the number of POI $m$. With a larger number of clusters, although there are some unused clusters of non-POI after clustering, the association step will sieve them and only choose the most consistent $m$ clusters with the event logs of $m$ device IDs. Lastly, for every image, its assigned ID is finalized by decision voting on the individual association results computed with different $g$, which will be soon introduced in the next paragraph. Typically, a majority voting procedure would give a hard ID label for an image and use these $<$Image, ID$>$ pairs as training data to update the face representation model. However, due to the noise in device presence observations and errors in clustering, there will be mis-labeling in the training data that may confuse the model update. To account for the uncertainty of ID assignment, we adopt soft labels for voting. Then instead of voting for the most likely ID $y_i$ for an image $x_i$, we introduce a probability vector $\mathbf{y}_{i} = (y_{i,1}, y_{i,2}, \ldots, y_{i,m})$. Specifically, every image is associated with all POI and its associated soft label is derived by the votes for the subjects divided by the total number of votes. In this way, the soft label of each image is a valid probability distribution that sums up to 1. For instance, $y_{i,j}=0.4$ means that there are $40\%$ associations assigning the ID $j$ to the image $i$. Moreover, the soft labeled data $<x_i, \mathbf{y}_{i}>$ is compatible with the computation of cross-entropy. 

\noindent \textbf{Number of Clusters}.
We are now in a position to describe how to select the number of clusters $g$. We observed that, with a proper clustering algorithm (i.e., agglomerative clustering in our choice), non-POI's images will form separate small clusters but they will not be chosen in our follow-up association step. This is because that the number of devices ID $m$ is always less than the number of clusters ($g \in [2*m, 5*m]$), and only those images clusters with the most consistent event attendances are associated. The small clusters of outlier images often have scatted attendance vector and will be ignored in association. Therefore, soft voting with varying number of clusters can reinforce the core images of POI by giving large probability and assigns non-POI's images with small probability.

\section{Model Updates} 
\label{sec:model_update}
We are now in a position to introduce the model update in {\sysname}. At every iteration, {\sysname} updates the last-iteration face representation model $f_{\theta}^{\tau}$ to $f_{\theta}^{\tau+1}$, by taking the labeled images as inputs. To correct the device observation errors, {\sysname} leverages these label images to update device attendance vector to $\mathbf{u}^{\tau}$ for all events. 

\subsection{Visual Model Update} 
\label{sub:visual_model_update}
\noindent \textbf{Discriminative Face Representation Learning}.
Face representation learning optimizes a representation loss $\mathcal{L}_R$ to enforce the learnt features as discriminative as possible. Strong discrimination bears two properties: inter-class dispersion and intra-class compactness. Inter-class dispersion pushes face images of different subjects away from one another and the intra-class compactness pulls the face images of the same subject together. Both of the properties are critical to face recognition. At iteration $\tau$, given the current labels $y_i^{\tau}$ for the $i$th face image $x_i$ and the transformed features $\mathbf{z}_i^{\tau} = f_{\theta}^{\tau}(x_i)$, the representation loss $\mathcal{L}_R$ is determined by a composition of softmax loss and center loss:
\begin{equation}
	\begin{split}
		\mathcal{L}_R &= \mathcal{L}_{softmax} + \mathcal{L}_{center} \\
				      &= \underbrace{\sum_{i}-log(\frac{e^{W_{y_i^{\tau}}^T\mathbf{z}_i^{\tau} + \mathbf{b}_{y_i^{\tau}}}}{\sum_{j=1}^{m}e^{W_{l_j}^T\mathbf{z}_i^{\tau} + \mathbf{b}_{l_j}}})}_\text{softmax loss} + \underbrace{\sum_{i}\frac{\lambda}{2} ||\mathbf{z}_i^{\tau} - \mathbf{o}_{y_i^{\tau}}||^2}_\text{center loss}
	\end{split}
	\label{equ:model_loss}
\end{equation}
where $W$ and $b$ are the weights and bias parameters in the last fully connected layer of the pre-trained model. $\mathbf{o}_{y_i^{\tau}}$ denotes a centroid feature vectors by averaging all feature vectors with the same identity label $y_i$. 
The center loss $\mathcal{L}_{center}$ explicitly enhances the intra-class compactness while the inter-class dispersion is implicitly strengthened by the softmax loss $\mathcal{L}_{softmax}$ \cite{wen2016discriminative}. $\lambda$ is a hyper-parameter that balances the above sub-losses. 

\noindent \textbf{Stochastic Center Loss}
The center loss $\mathcal{L}_{center}$ in Eq.~\ref{equ:model_loss} is shown to be helpful to enhance the intra-class compactness \cite{wen2016discriminative}. However, we cannot directly adopt it for fine-tuning as computing the centers requires explicit labels (see Eq.~\ref{equ:model_loss}) of images, while the association steps above only provide probabilistic ones through soft labels. To solve this, we propose a new loss called \emph{stochastic center loss} $\mathcal{L}_{stoc}$ to replace the center loss. Similar to the idea of fuzzy sets \cite{xie1991validity}, we allow each face image to belong to more than one subject. The membership grades indicate the degree to which an image belongs to each subject and can be directly retrieved from the soft labels and the stochastic center $\mathbf{o}_k^{\tau}$ for the $k$-th identity is given as:

\begin{equation}
	\mathbf{o}_k^{\tau} = \frac{\sum_i^{n}\mathbf{z}_i^{\tau} * y_{i,k}^{\tau}}{\sum_i^{n} y_{i,k}^{\tau}}
	\label{equ:fuzzy_center}
\end{equation}
This gives the stochastic center loss as follows:
\begin{equation}
\mathcal{L}_{stoc} = \sum_{i}\sum_{k} y_{i,k}^{\tau} *||\mathbf{z}_i^{\tau} - \mathbf{o}_k^{\tau}||^2
\end{equation}

We leave the softmax loss $\mathcal{L}_{softmax}$ the same as in Eq.~\ref{equ:model_loss}, because the soft labels are compatible with the computation of cross-entropy. The new representation loss to minimize is:
\begin{equation}
	\mathcal{L}_R = \mathcal{L}_{softmax} + \mathcal{L}_{stoc}
\end{equation}
{\sysname} updates the model parameters $\theta^{\tau}$ to $\theta^{\tau+1}$ based on the gradients of $\nabla_{\theta}\mathcal{L}_R$, which are calculated via back prorogation of errors. Compared with the dataset used for pre-training, which is usually in the order of millions \cite{cao2017vggface2}, the data used for fine-tuning is much smaller (several thousands). The mis-match between the small training data and the complex model architecture could result in overfitting. To prevent overfitting, we use the dropout mechanism in training, which is widely adopted to avoid overfitting \cite{lecun2015deep}. Meanwhile, as observed in \cite{hinton2015distilling,reed2014training}, the soft label itself can play the role of a regularizer, and make the trained model robust to noise and reduce overfitting. Fig.~\ref{fig:model_update} illustrates the effect of model update.

\subsection{User Attendance Update} 
\label{sub:user_attendance_update}
The device presence observations by WiFi sniffing are noisy because the WiFi signal of a device is opportunistic and people do not carry/use theirs devices all the time. Based on the results of the cluster labelling step introduced in \sect\ref{sub:cluster_labeling}, we have the opportunity to update our belief on which users attended each event. The update mechanism is as follows: Each image is associated with a user probability vector, whose elements denote the probability that the image correspondences to a particular user. By averaging the user probability vectors of all images that have been drawn from the same event ${e_k}$, and normalizing the result, we can estimate the user attendance of this event. The elements of the resulting user attendance vector $\widehat{\mathbf{u}}_k^{\tau}$ denote the probabilities of different users attending event $e_k$. 

We can now use $\widehat{\mathbf{u}}_k^{\tau}$ as a correction to update our previous wifi attendance vector $\mathbf{u}_{k}^{\tau}$  as follows:

\begin{equation}
	\mathbf{u}^{\tau+1}_{k} = \mathbf{u}^{\tau}_{k} - \gamma \cdot (\mathbf{u}^{\tau}_{k} - \widehat{\mathbf{u}}_k^{\tau})
\end{equation}
where $\gamma$ is a pre-defined parameter that controls the ID update rate. In principle, a large update rate will speed up the convergence rate, at the risk of missing the optima. {\sysname} sequentially repeats the above steps of clustering, labelling and model updates, until the changes $\xi$ in the user attendance model are negligible ($\leq 0.01$ in our case). Algorithm.~\ref{alg:autotune} summarizes the workflow.

\begin{algorithm}[t] 
\KwIn{pre-trained model $f_{\theta}^{1}$, images $\mathcal{X}$, POI's device IDs $\mathcal{L}$, threshold $\xi$}
\KwOut{adapted model $f_{\theta}^{\ast}$, corrected attendance observations $I^{\ast}$, soft image labels $Y^{\ast}$} 
\kwInit{Given sniffed $\mathcal{L}$ in all events $\mathcal{E}$, compute attendance vector $\mathbf{u}^0$}
$\tau=1$ \\
\While{$\sqrt{\frac{1}{h}\sum_{k=1}^{h}||\mathbf{u}^{(\tau)}_{k} - \mathbf{u}^{(\tau-1)}_{k}||^2} > \xi$}{
	$\mathcal{Z}^\tau = f_{\theta}^{\tau}(\mathcal{X})$ \Comment{feature transformation}\\
	\For{$g \leftarrow 2*m$ to $5*m$}{
	 	$\mathcal{C}^\tau \leftarrow$ \texttt{cross\_modality\_clustering}($\mathcal{Z}^\tau$, $\mathbf{u}^\tau$, g) \Comment{\sect\ref{sub:cross_modality_clustering}}\; 
	 	$A^\tau_g \leftarrow$ \texttt{cluster\_labeling}($\mathcal{C}^\tau$, $\mathbf{r}_c^\tau$, $\mathbf{r}_l^\tau$) \Comment{\sect\ref{sub:cluster_labeling}}\; 
	}
	$\mathcal{Y}^{\tau} \leftarrow$ \texttt{soft\_voting}($A^\tau$) \Comment{\sect\ref{sub:cluster_labeling}}\;
 	$\mathbf{o}^{\tau} \leftarrow$ \texttt{stochastic\_center}($\mathcal{Y}^{\tau}$, $\mathcal{Z}^\tau$) \Comment{\sect\ref{sub:visual_model_update}}\; 
 	$f_\theta^{\tau+1} \leftarrow$ \texttt{visual\_model\_update}($\mathcal{X}$, $\mathcal{Y}^\tau$, $\mathbf{o}^{\tau}$, $f_\theta^{\tau}$) \Comment{\sect\ref{sub:visual_model_update}} \;
 	$\mathbf{u}^{\tau+1} \leftarrow$ \texttt{user\_attendance\_update}($\mathcal{Y}^{\tau}$, $\mathbf{u}^{\tau}$) \Comment{\sect\ref{sub:user_attendance_update}} \;
  $\tau \leftarrow \tau + 1$
 }
    \caption{{\bf {\sysname}} \label{alg:autotune}}

\end{algorithm}

\section{Implementation} 
\label{sec:implementation}
In this section, we introduce the implementation details of {\sysname} (code available at \url{https://github.com/Wayfear/Autotune}).

\subsection{Heterogeneous Data Sensing} 
\label{sub:heterogeneous_data_sensing}

\noindent \textbf{Face Extraction}.
This module consists of a front-end remote camera and a back-end computation server~\footnote{The study has received ethical approval R50950}. Specifically, the remote cameras in our experiment are diverse and include \emph{GoPro Hero 4}\footnote{\url{https://shop.gopro.com/cameras}}, \emph{Mi Smart Camera}\footnote{\url{https://www.mi.com/us/mj-panorama-camera/}} and \emph{Raspberry Pi Camera}\footnote{\url{https://www.raspberrypi.org/products/camera-module-v2/}}). We modified these cameras so that they are able to communicate and transfer data to the back-end through a wireless network. To avoid capturing excess data without people in it, we consider a motion-triggered mechnism with a circular buffer. It works by continuously taking low-resolution images, and comparing them to one another for changes caused by something moving in the camera's field of view. When a change is detected, the camera takes a higher-resolution video for $5$ seconds and reverts to low resolution capturing. All the collected videos are sent to the backend at every midnight. On the backend, a cascaded convolutional network based face detection module \cite{zhang2016joint} is used to drop videos with no face in them. The cropped faces from the remaining videos are supplied to {\sysname}.

\noindent \textbf{WiFi Sniffing}.
This module is realized on a WiFi-enabled laptop running Ubuntu 14.04. Our sniffer uses Aircrack-ng \footnote{\url{https://www.aircrack-ng.org/}} and tshark \footnote{\url{https://www.wireshark.org/docs/man-pages/tshark.html}} to opportunistically capture the WiFi packets in the vicinity. The captured packet has unencrypted information such as transmission time, source MAC address and the Received Signal Strengths (RSS). As {\sysname} aims to label face images for POI, our WiFi sniffer only records the packets containing MAC addresses of POI's and discards them otherwise, so as to not harvest addresses from people who have not given consent. A channel hop mechanism is used in the sniffing module to cope with cases where the POI's device(s) may connect to different WiFi networks, namely, on different wireless channels. The channel hop mechanism forces the sniffing channel to change by every second and monitor the active channels periodically (1 second) in the environment. The RSS value in the packet implies how far away the sniffed device is from the sniffer \cite{lu2016robust,lu2016elm}. By putting the sniffer near the camera, we can use a threshold to filter out those devices with low RSS values, e.g., less than -55 dBm in this work, as they are empirically unlikely to be within the camera's field of view.

\noindent \textbf{Event Segmentation}.
Depending on the context, the duration of events can be variable. However, for simplicity we use fix-duration events in this work. Specifically, we split a day into 12 intervals, each of which is $2$ hours long. We then discard those events that have neither face images nor sniffed MAC addresses of POI.

\subsection{Face Recognition Model} 
\label{sub:face_recognition_model}
The face recogonition model used in {\sysname} is the state-of-the-art FaceNet \cite{schroff2015facenet}.

\noindent \textbf{Pre-training}.
FaceNet adopts the Inception-ResNet-v1 \cite{szegedy2017inception} as its backbone and its weights are pre-trained on the VGGFace2 dataset \cite{cao2017vggface2}. This dataset contains $3.31$ million
images of $9131$ subjects, with an average of $362.6$ images for each subject. Images are downloaded from Google Image Search and have large variations in pose, age, illumination, ethnicity and profession.
Pre-training is supervised by the triplet loss \cite{cheng2016person} and the training protocols, e.g, parameter settings, can be found in \cite{schroff2015facenet}. We found that the learnt face representation by FaceNet is generalizable and it is able to achieve an accuracy of 99.65\% on the LFW face verification task \footnote{\url{http://vis-www.cs.umass.edu/lfw/}}. Note that, FaceNet \cite{schroff2015facenet} not only learns a powerful recognition model but gives a very discriminative feature representation that can be used for clustering.

\noindent \textbf{Fine-tuning by \sysname}. The fine-tuning process has been explained in \sect\ref{sub:visual_model_update}, and here we provide some key implementation detials. After each round of label association (see \sect\ref{sub:cluster_labeling}), the labeled data is split into a training set and validation set, with a ratio of $8:2$ respectively. The pre-trained FaceNet is then fine-tuned on the training set, and the model that achieves the best performance on the validation set is saved. Note that the fine-tuning process in {\sysname} does not involve the test set. The online testing is performed on a held-out set that is collected on different days. To enhance the generalization ability, we use dropout training for regularization \cite{wager2013dropout}. The dropout ratio is set to $0.2$. We set the batch size to $50$ and one fine-tuning takes $100$ epochs. 

\subsection{System Configuration} 
\label{sub:parameter_configuration}

\noindent \textbf{Face Detection}.
As discussed in \sect\ref{sub:heterogeneous_data_sensing}, we use a cascaded convolution network to detect faces in videos. It is cascaded by three sub-networks, a proposal network, a refine network and an output network. Each of them can output bounding boxes of potential faces and the corresponding detection probabilities, i.e., confidences. Face detection with small confidence will be discarded early in the process and not sent to the next sub-network. In this work, the confidence threshold is set to $0.7$, $0.7$ and $0.9$ for three sub-networks respectively. Following the original setting in \cite{zhang2016joint}, we set the minimal face size in detection to $40 \times 40$ pixels.

\noindent \textbf{Setup of Face Clustering}.
In \sect\ref{sub:cross_modality_clustering}, we use a clustering algorithm to merge the images across events. Specifically, the clustering algorithm used in {\sysname} is agglomerative clustering \cite{beeferman2000agglomerative}, which is a method that recursively merges the pair of clusters that minimally increases a given linkage distance. The similarity metric adopted here is Euclidean distance. A linkage criterion determines which distance to use between sets of data points. In {\sysname}, this linkage criterion is set to the average distances of all sample pairs from two sets. \chris{As introduced in Sec.~\ref{sub:cluster_labeling}, the number of clusters $g$ is determined by the number of POI $m$. We vary $g$ from $2*m$ to $5*m$ and proceed soft voting, to account for the extra clusters of non-POI's face images and ambiguous/outlier images of POI.}

\section{Evaluation} 
\label{sec:evaluation}
In this section, we evaluate the \sysname extensively on datasets collected from both real-world experiments and simulation. We deployed two testbeds, one in the UK and the other in China, and collected datasets as described in the previous section. The simulation dataset is developed based on a public face datasets. 

\subsection{Evaluation Protocols} 
\label{sub:competing_approaches_and_evaluation_metrics}

\noindent \textbf{Competing Approaches}.
We compare the performance of \sysname with the $3$ competing approaches:
\begin{itemize}
  \item \textbf{Template Matching (TM)} \cite{best2014unconstrained} employs a template matching method to assign ID labels to clusters of face images. This is the most straightforward method that is used when one or more profile photos of POI are available, e.g., crawled from their personal homepage or Facebook.
  \item \textbf{One-off Association (OA)} \cite{lu2017towards} uses one-off associations to directly label the image clusters without fine-tuning of the face representation model itself.
  \item \textbf{Deterministic \sysname (D-AutoTune)} is the deterministic version of \sysname. In D-AutoTune, the association and update steps are the same as in \sysname, but it adopts hard labels rather than soft labels and uses the simple center loss instead of the proposed stochastic center loss (see \sect\ref{sub:visual_model_update}).
\end{itemize}

\noindent \textbf{Evaluation Metrics}.
AutoTune contains two main components, offline label assignment and online inference. For the offline label assignment, we evaluate its performance with the following metrics: TP, TN, FP, FN are true positive, true negative, false positive and false negative respectively. Each metric captures different aspects of classification \cite{powers2011evaluation}. Online face recognition has two kinds of tests, face identification and verification. We follow \cite{liu2017sphereface,liu2017deep} to use Cumulative Match Characteristic (CMC) for the evaluation of online face identification.

\begin{table*}[!t]
\centering
\small
\caption{Key Metrics of Collected Dataset Sets.}
\label{tab:stats_tuning}
\begin{tabular}{|c|c|c|c|c|c|c|c|l|c|}
\hline
\textbf{Site} & \multicolumn{1}{l|}{\textbf{\begin{tabular}[c]{@{}l@{}}NO. of\\ Rooms\end{tabular}}} & \textbf{\begin{tabular}[c]{@{}c@{}}NO. of \\ POI\end{tabular}} & \textbf{\begin{tabular}[c]{@{}c@{}}NO. of \\ non-POI\end{tabular}} & \textbf{\begin{tabular}[c]{@{}c@{}}NO. of\\ Images\end{tabular}} & \textbf{\begin{tabular}[c]{@{}c@{}}NO. of\\ Events\end{tabular}} & \textbf{\begin{tabular}[c]{@{}c@{}}Event\\ Duration\end{tabular}} & \textbf{\begin{tabular}[c]{@{}c@{}}Average\\ subjects/event\end{tabular}} & \multicolumn{1}{c|}{\textbf{\begin{tabular}[c]{@{}c@{}}Camera(s)\\ in rooms\end{tabular}}} & \textbf{\begin{tabular}[c]{@{}c@{}}Experiment\\ Note\end{tabular}} \\ \hline
\textbf{UK} & 3 & 24 & 11 & 15, 286 & 83 & 3h & 9.14 & \begin{tabular}[c]{@{}l@{}}Office: GoPro Hero 4\\ Kitchen: Pi Cam \& Mi Cam\\ Meeting: Pi Cam \& Mi Cam\end{tabular} & \begin{tabular}[c]{@{}c@{}}GoPro: 1080p, 90FPS\\ PiCam: 720p, 90FPS\\ MiCam: 720p, 15FPS\end{tabular} \\ \hline
\textbf{CHN} & 1 & 12 & 25 & 7, 495 & 102 & 2h & 3.36 & CommonRM: Gopro Hero 4 & All Faces are from Asian \\ \hline
\end{tabular}
\end{table*}

\begin{figure*}[t!]
	\centering
	\begin{minipage}{0.32\textwidth}
		\centering
		\begin{subfigure}{0.98\textwidth}
			\includegraphics[width=\linewidth]{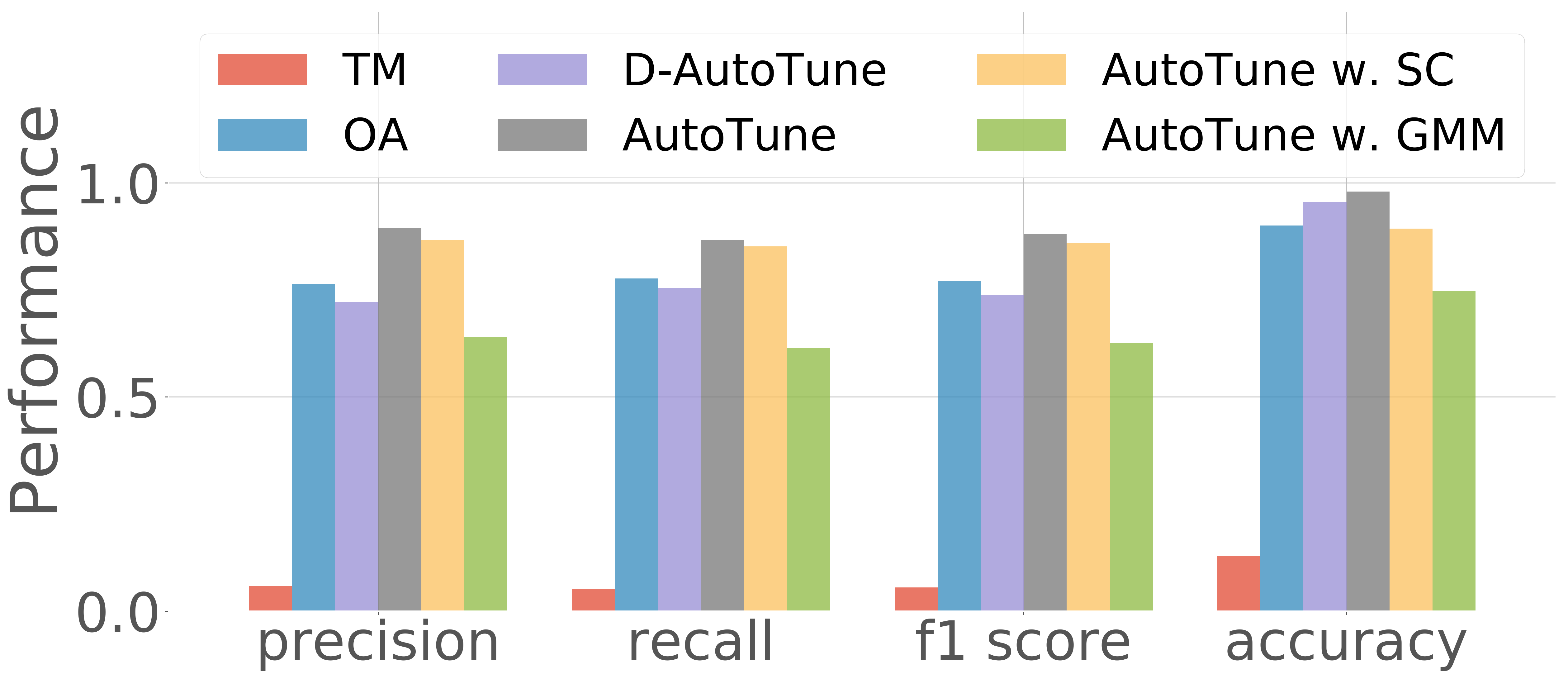}
			\caption{Office, UK} \label{fig:assoc_oxford}
		\end{subfigure}
		\newline 
		\noindent 
		\begin{subfigure}{0.98\textwidth}
			\includegraphics[width=\linewidth]{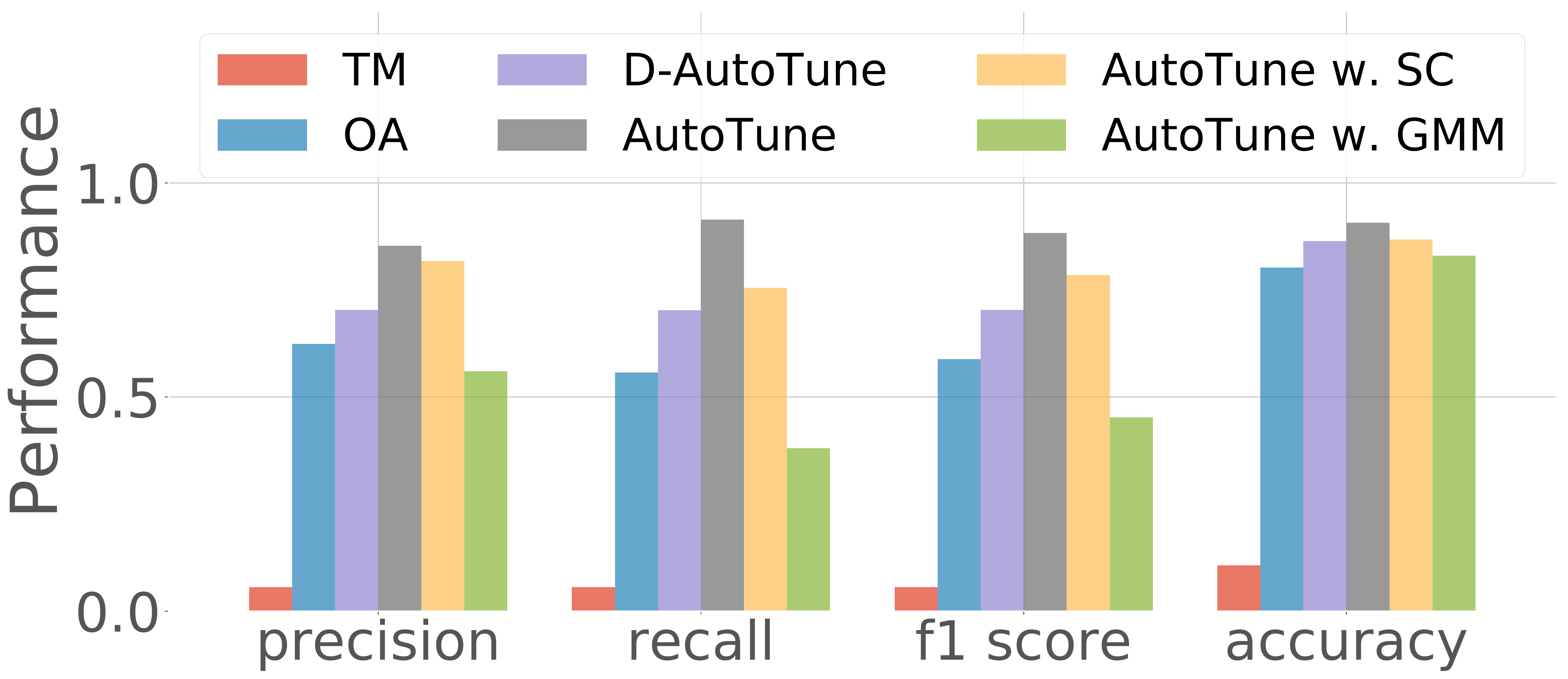}
			\caption{CommonRoom, CHN} \label{fig:assoc_tongji}
		\end{subfigure}
		\caption{Overall Labeling Performance.} \label{fig:result_assoc}
	\end{minipage}
	\begin{minipage}{0.33\textwidth}
		\centering
		\begin{subfigure}{\textwidth}
		\includegraphics[width=\linewidth]{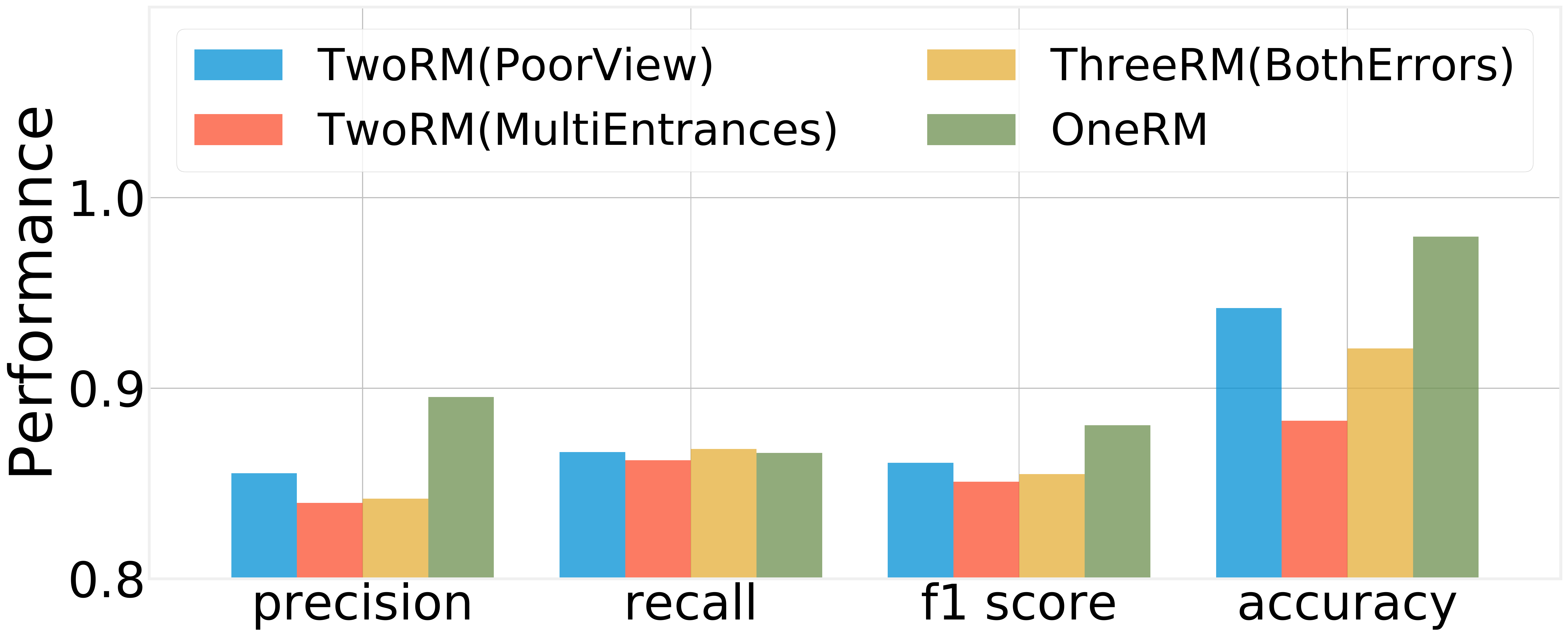}
		\caption{Raspberry Pi Camera} \label{fig:pi_cam_tuning}
		\end{subfigure}
		\newline 
		\noindent 
		\begin{subfigure}{\textwidth}
		\includegraphics[width=\linewidth]{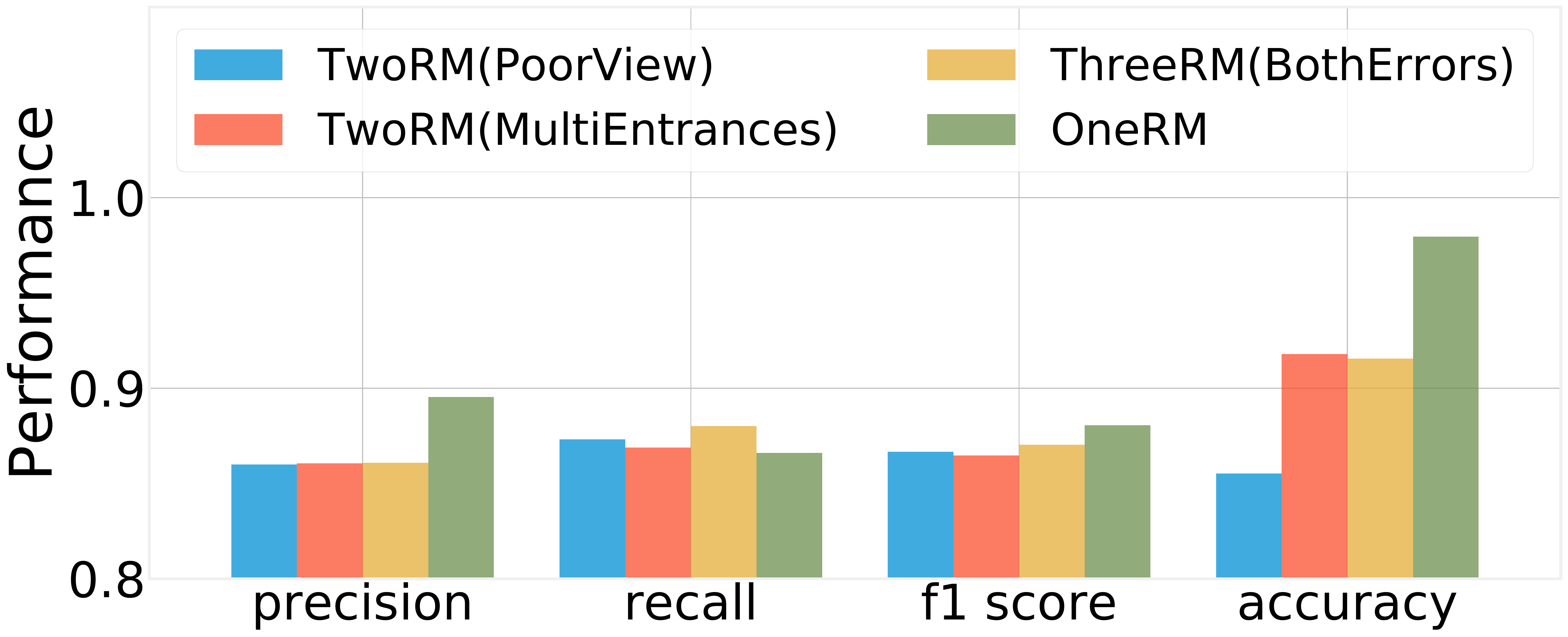}
		\caption{Mi Smart Camera} \label{fig:mi_cam_tuning}
		\end{subfigure}
		\caption{Performance of Scalability.} \label{fig:cross_camera_tuning}
	\end{minipage}
		\begin{minipage}{0.33\textwidth}
			\begin{subfigure}{\textwidth}
			\includegraphics[width=\linewidth]{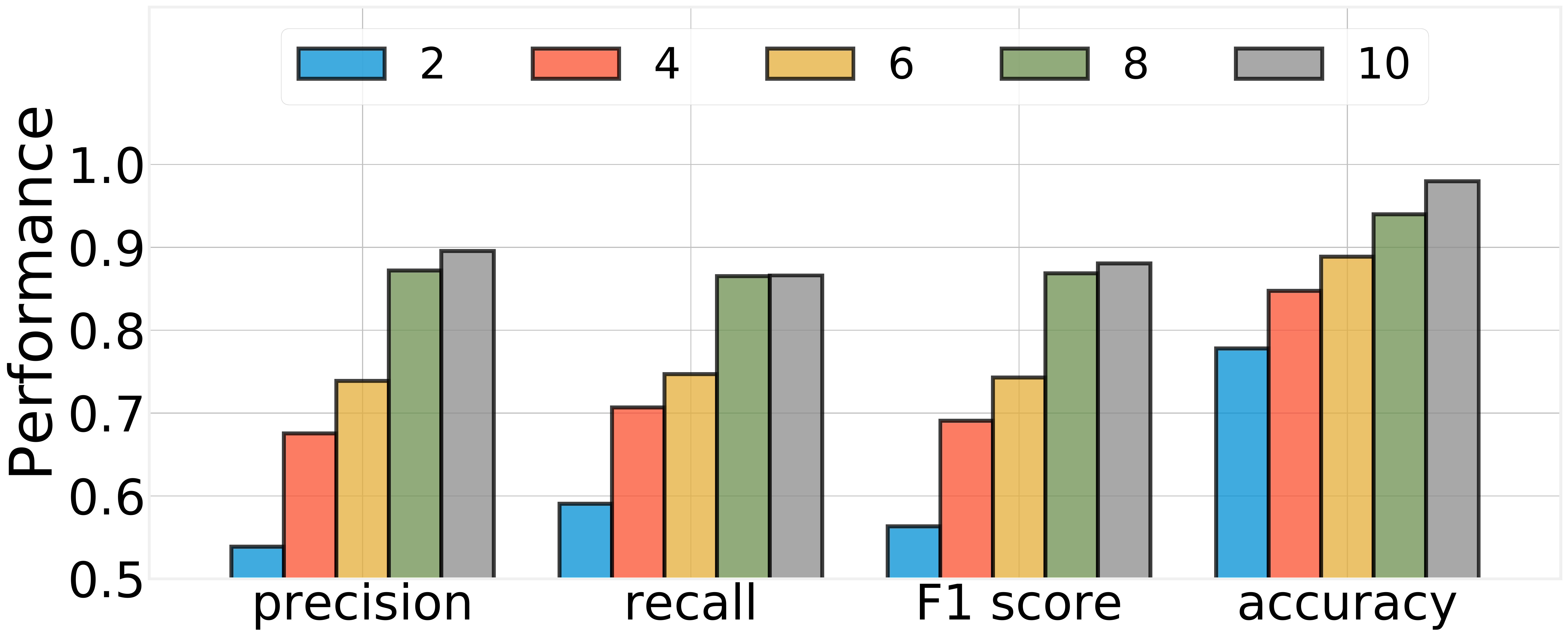}
			\caption{UK} \label{fig:span_oxford}
			\end{subfigure}

			\begin{subfigure}{\textwidth}
			\includegraphics[width=\linewidth]{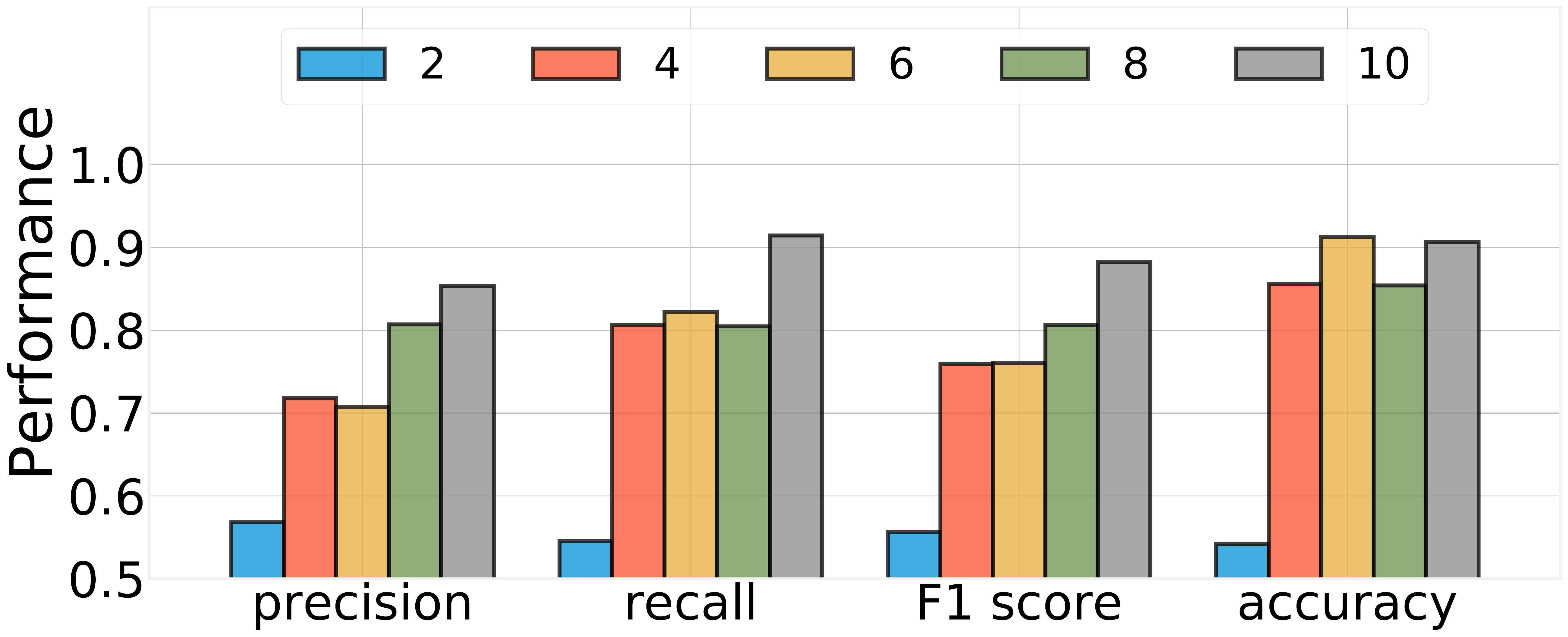}
			\caption{CHN} \label{fig:span_tongji}
			\end{subfigure}
		\caption{Impact of Number of Days.} \label{fig:result_span}
	\end{minipage}
\end{figure*}

\subsection{Offline Cross-modality Face Labeling} 
\label{sub:offline_cross_modality_face_labeling}
AutoTune automatically labels images captured in the wild by exploiting their correlations with the device presences. The quality of image labeling is crucial for the follow-up face recognition in the online stage. In this section, we investigate the image labeling performance of \sysname.

\subsubsection{Data Collection} 
\label{ssub:data_collection}
We deployed {\sysname} at the testbeds in two countries, with different challenging aspects.

	\noindent \textbf{UK Site:} 
	This first dataset is collected in a commercial building in the UK. We deploy the heterogeneous sensing front-ends, including surveillance cameras and WiFi sniffers, on a floor with three different types of rooms: office, meeting room and kitchen. $24$ long-term occupants work inside can freely transit across these rooms. These occupants are naturally chosen as people of interest (POI). For the office, face images are captured with a surveillance camera that faces the entrance. The presence logs of occupants' WiFi MAC addresses are collected by a sniffer that is situated in the center of the room for the same time period. Besides the POI's faces, these images also contain the faces of $11$ short-terms visitors who came to this floor during the experiments. \eat{Along with the face images, the presence logs of occupants WiFi MAC addresses are also collected for the same time period.} 
	We put different cameras in different rooms to examine the performance of \sysname under camera heterogeneity. To further examine the resilience of \sysname, we put cameras in adversarial positions. In kitchen, we deploy cameras with bad views near entrance so that they can only capture subjects above 1.7m. While in the meeting room with two entrances, only the primary entrance is equipped with cameras. Therefore in both rooms cameras constantly \emph{mis-capture} face images of subjects. Tab.~\ref{tab:stats_tuning} summarizes this data collection.

	\noindent \textbf{CHN Site:}
	We collect another dataset in a common room of a university in China. There are no long-term occupants in this site and all undergraduates can enter. Of the $37$ people that appeared during the three week period, $12$ subjects are selected as the POI, and their WiFi MAC address presence is continuously recorded by the sniffer. Other settings remain the same as the UK site. The challenge in this dataset lies in that the captured face images, both for POI and non-POI, are all of Asian people, while the initial face representation model is trained primarily on Caucasians. Details of the CHN dataset are given in Tab.~\ref{tab:stats_tuning}. Observation noises are very common in this dataset.    
	
\subsubsection{Overall Labeling Performance} 
\label{ssub:overall_labeling_performance}
We start our evaluation with one room only and compare our results with baselines. Fig.~\ref{fig:result_assoc} shows the performance of label assignment, i.e., matching an identifier to a face image. For the office dataset, \sysname outperforms the best competing approach (OA), by $0.13$ in $F_1$ score and $7\%$ in accuracy. The advantage of \sysname is more obvious in the CommonRoom experiment where it beats the best competing approach (OA) by $0.34$ in $F_1$ score and $10\%$ in accuracy. As the only method that uses the website images (one-shot learning) to label images rather than the device ID information, TM struggles in both experiments and is $9$-fold worse than \sysname. We observe that the website face images are dramatically different from the captured images in real-world, due to different shooting conditions and image quality. These results imply that, although the device observations via WiFi sniffing are noisy, when the amount of the them is enough, they are more informative than the web-crawled face images.

Additionally, we note that adopting soft labels (see \sect\ref{sub:cluster_labeling}) can further improve the labeling performance. In terms of $F_1$ score, the full-suite \sysname is around $15\%$ and $22\%$ better than D-AutoTune in the two experiments respectively. Similar improvements are witnessed in terms of the accuracy. The reason of the larger performance gap in the CommonRoom dataset is that there are more images of non-POI being captured in this experiment. In addition, the pre-trained FaceNet model does not generalize very well to Asian faces, which are the primary ethnicity in the CommonRoom.

Lastly, we found that the choice of algorithm for cross-event clustering affects the final label-association accuracy. The best performance is achieved with the default agglomerative clustering algorithm on both datasets. The second-best spectral clustering (SC) is slightly inferior to agglomerative clustering in the office dataset, though the gap between them gets large in the CommonRoom dataset. The Gaussian Mixture Model (GMM) is inferior to the other two clustering algorithms. This is because GMM is best suited to Mahalanobis-distance based clustering whereas the distance space defined in Eq.~\ref{equ:id-ass_sim}  is a non-flat space due to the introduced attendance information. Hence, its resultant clusters are very impure and give poor association performance.


\subsubsection{Performance vs. Scalability} 
\label{ssub:performance_of_robustness}
We further examine {\sysname} with multiple rooms under different adversarial conditions. 
Together with the \textit{UK office} data, we evaluate \sysname when events are collected in three different locations and via heterogeneous cameras. In particular, a kitchen and a meeting room are included and each of them has two lower-fidelity cameras (Mi Home Camera and Raspberry Pi Camera). The setup of multi-location experiments is given in Tab.~\ref{tab:stats_tuning}. As described in \sect{\ref{ssub:data_collection}}, there are many face mis-detections in the kitchen and meeting room due to the adversarial camera setups. Nevertheless, as shown in Fig.~\ref{fig:cross_camera_tuning}, \sysname only suffers little performance drop ($\leq 0.03$) when erroneous camera observations are mixed with the single-office data. In terms of F1 score, which is the most important metric, \sysname can achieve comparable level with the office experiment, regardless of which camera is added. Moreover, \sysname still maintains its good performance even in the most adversarial case, where both errors are mixed in (three-room experiment). The main reason is that by using more data, though noisy, extra validation constraints are also utilized by \sysname and make itself robust to observation errors.

\begin{figure}[t!]
	\centering
	\begin{subfigure}{0.23\textwidth}
	\includegraphics[width=\linewidth]{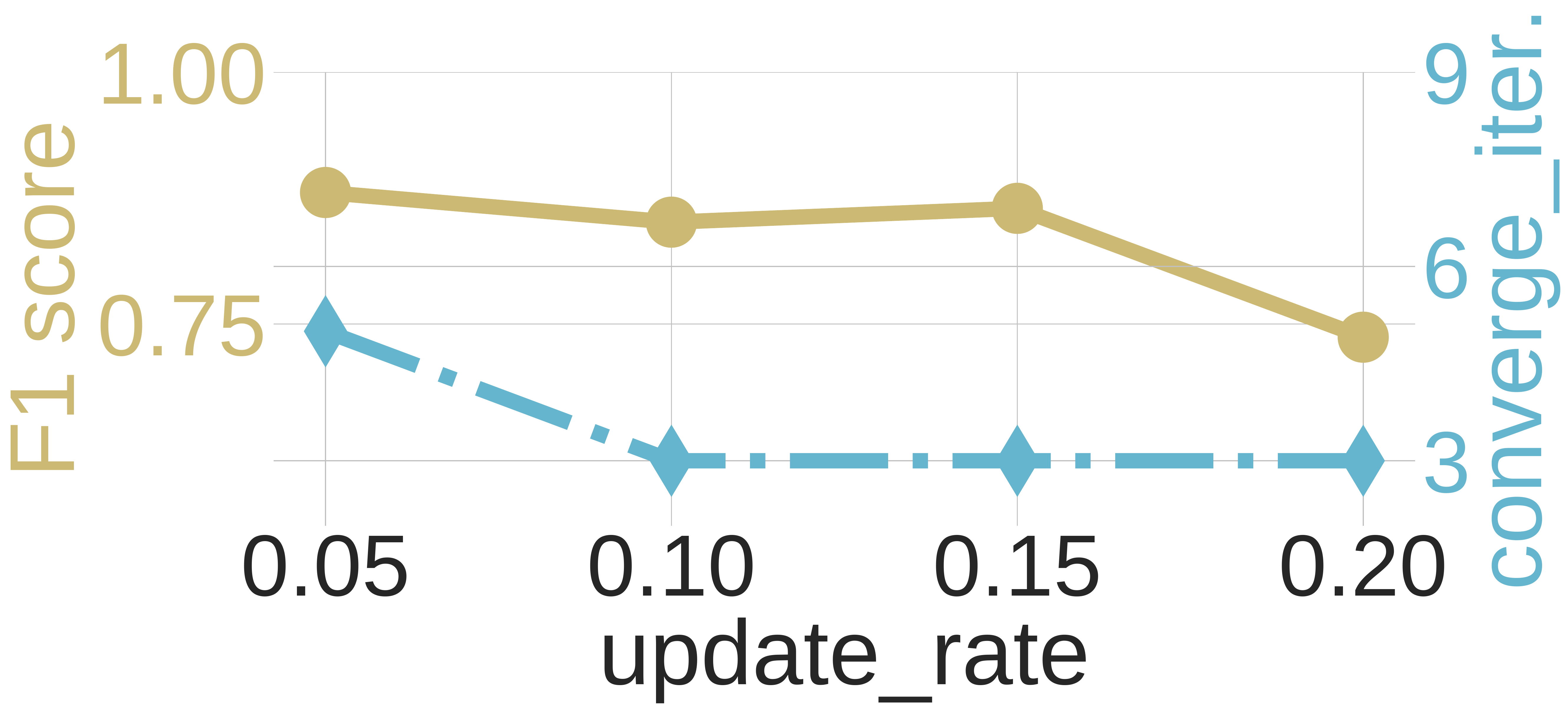}
	\caption{UK} \label{fig:error_rate_oxford}
	\end{subfigure}
	\hspace*{\fill}
	\begin{subfigure}{0.23\textwidth}
	\includegraphics[width=\linewidth]{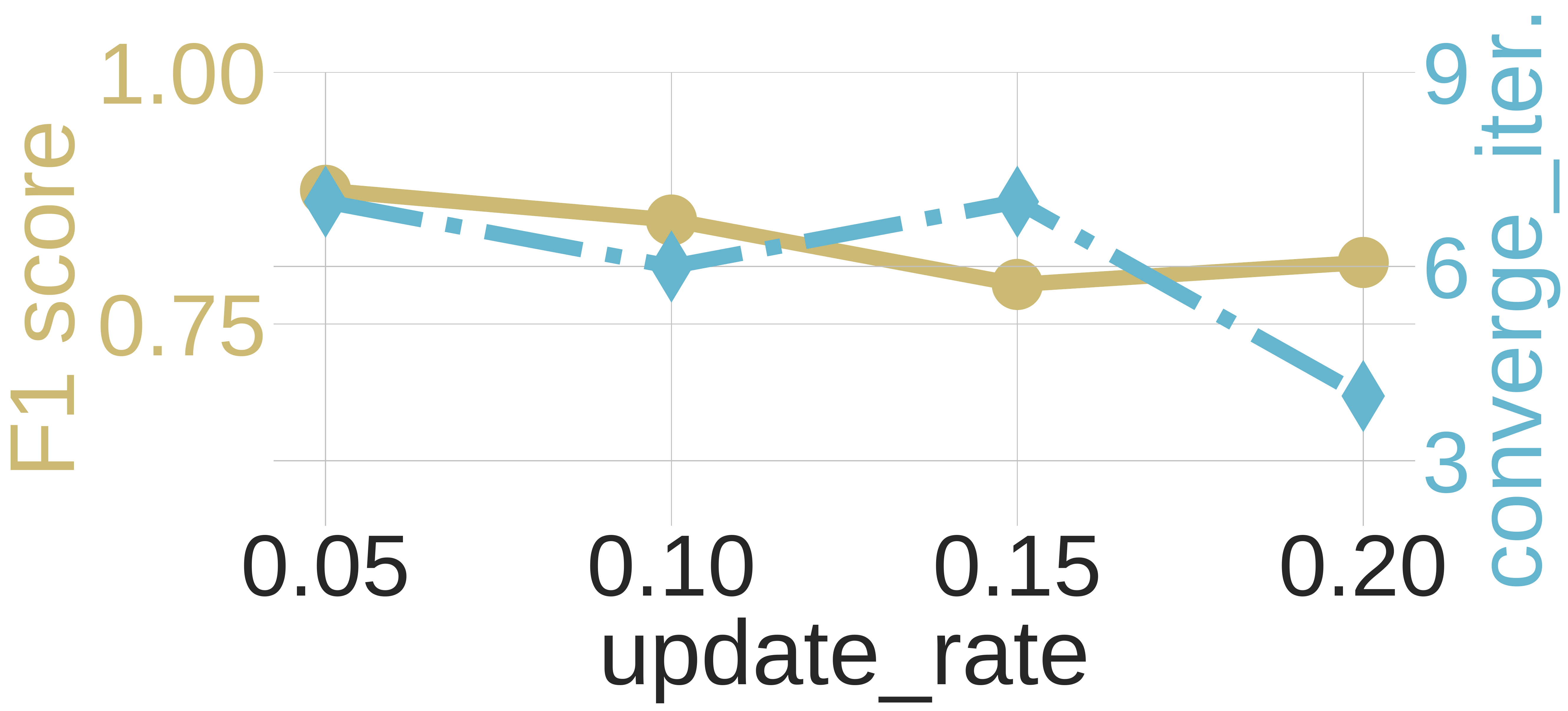}
	\caption{CHN} \label{fig:error_rate_tongji}
	\end{subfigure}
\caption{Impact of ID update rate $\beta$ (introduced in \sect\ref{sub:user_attendance_update}). It can be seen that a high update rate causes the network to converge rapidly to potentially incorrect assignments.} \label{fig:result_error_rate}
\end{figure}

\subsubsection{Performance vs. Lifespan} 
\label{ssub:impact_of_number_of_events}
{\sysname} exploits co-located device and face presence in different events to establish the cross-modality correlations. In this section, we investigate the impact of the collection span on the performance of labeling. Longer collection days give more events. We investigate its impact by feeding \sysname with data collected in different number of days, and compare them with all days on two datasets respectively. Fig.~\ref{fig:result_span} shows that \sysname performs better with increasing number of days on both datasets. The gap of $F_1$ score between the case with all days ($10$ days) and case with the least amount of days ($2$ days) can be as large as $>0.32$ on both datasets. As discussed in \sect\ref{sub:cluster_labeling}, the ID association needs sufficiently diverse events to create discriminative enough event vectors. Otherwise, there will be faces or devices with the same event vectors that hinders \sysname's ability to disambiguate their mutual association. However, we also observe that when we collect more than $8$ days, the performance improvement of \sysname becomes marginal. 

\subsubsection{Impact of ID update rate $\beta$.} 
\label{ssub:impact_of_id_update_rate_beta_}
This section investigates the impact of ID update rate $\beta$ introduced in \sect\ref{sub:user_attendance_update}. The ID update uses the fine-tuned face recognition model to update the device ID observations. A large ID update rate forces the device ID observations to quickly become consistent with the deep face model predictions. However, a large update rate also runs the risk of missing the optima. We vary the update rate $\beta$ from $0.05$ to $0.20$ at a step length of $0.05$. Fig.~\ref{fig:error_rate_oxford} demonstrates that \sysname achieves the best performance on the UK dataset when the update rate is set to $0.05$. The performance declines by $10\%$ when the rate rises to $0.2$. This is because the updated ID observations quickly become the same as the model predictions and the model	predictions are not quite correct yet. When it comes to CHN dataset (see Fig.~\ref{fig:error_rate_tongji}), similar trend of $F_1$ score change can be seen. Although overall, the convergence becomes faster when the update increases, we observe that there is a fluctuation point at the update rate of $0.15$, where \sysname takes $7$ iterations to converge. By inspecting the optimization process, we found that, under this parameter setting, \sysname oscillated because the large update step makes it jump around in the vicinity of the optima but it is unable to approach it furthermore. In practice, we suggest users of \sysname to select their update rate from a relatively safe region between $0.05$ to $0.1$.

\begin{figure}[t!]
	\centering
	\begin{subfigure}{0.23\textwidth}
	\includegraphics[width=\linewidth]{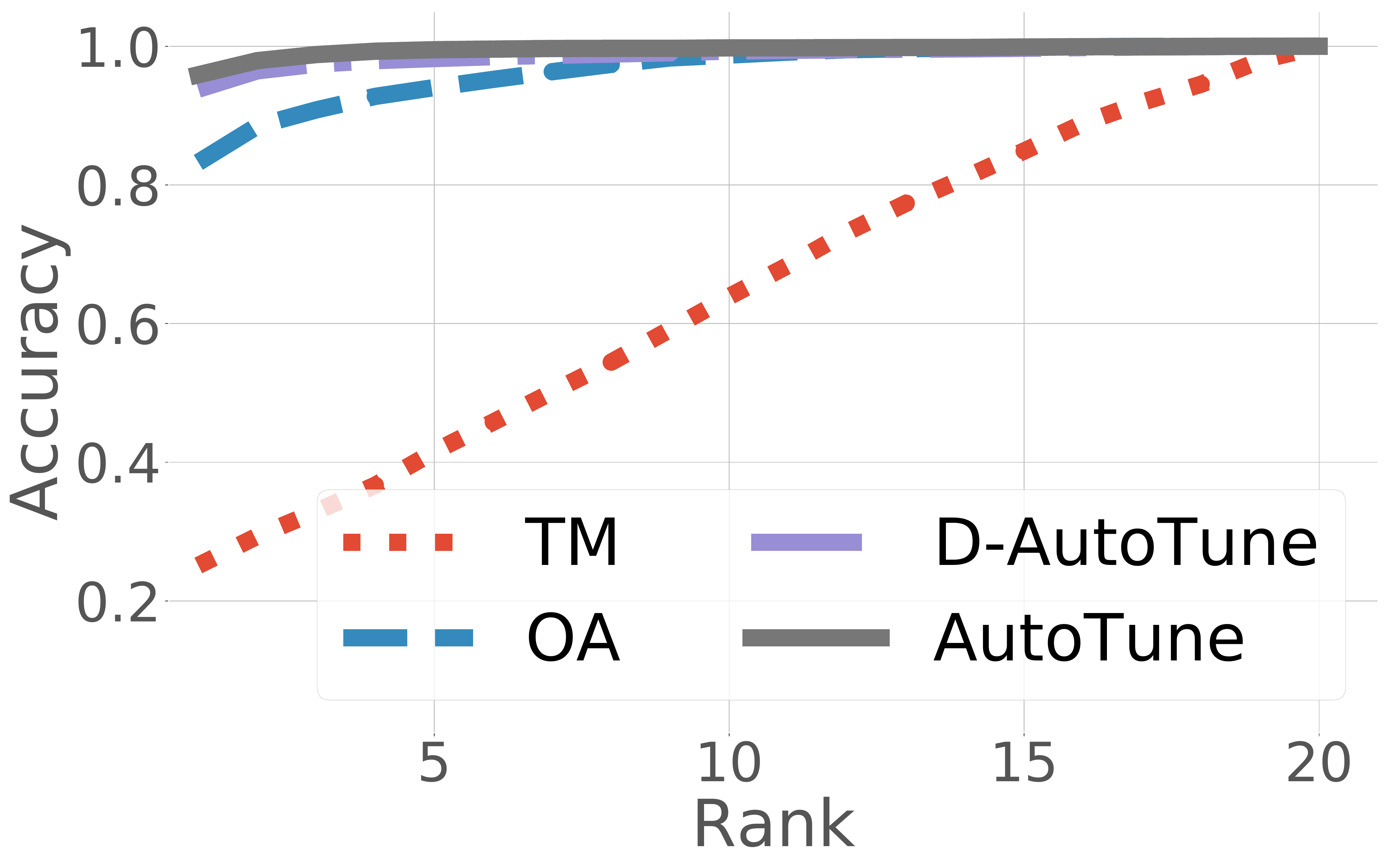}
	\caption{UK} \label{fig:cmc_oxford}
	\end{subfigure}
	\hspace*{\fill}
	\begin{subfigure}{0.23\textwidth}
	\includegraphics[width=\linewidth]{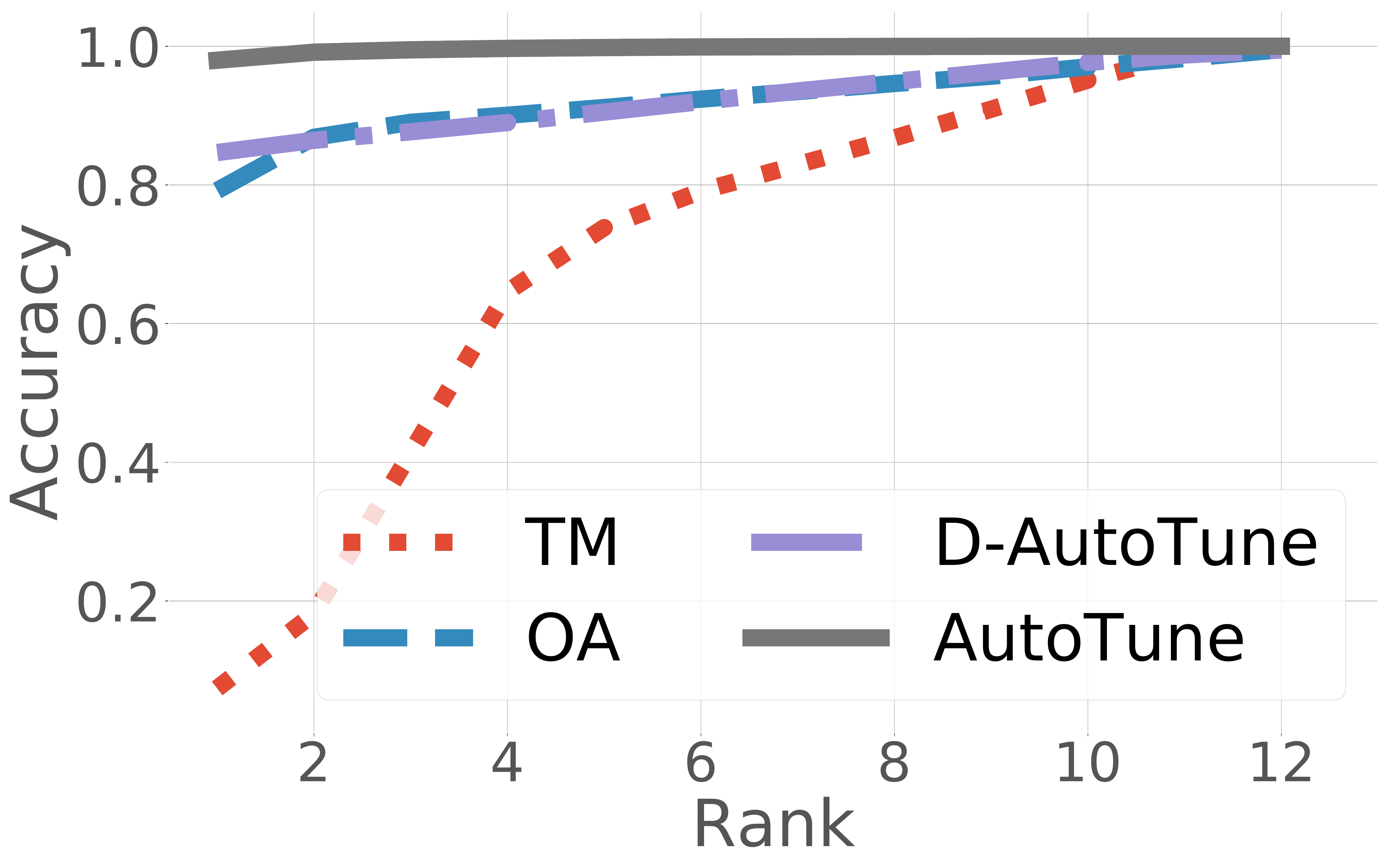}
	\caption{CHN} \label{fig:cmc_tongji}
	\end{subfigure}
	\caption{Online Identification Performance} \label{fig:identification}
\end{figure}

\subsection{Online Face Identification} 
\label{sub:online_face_recognition}
\subsubsection{Setup} 
\label{ssub:setup}
After the cross-modality label association, an image database of POI is developed and an online face recognition system can be trained based on it. As in \cite{schroff2015facenet}, we use the face representation model to extract face features and then use a classifier, say linear SVM in this work, to recognize faces. Particularly, as \sysname and D-AutoTune are able to update the face representation model in their labeling process, the feature extractor for them are the updated model by themselves. For competing approaches that do not have an evolving model updates, we use the original Facenet model (in \sect\ref{sub:face_recognition_model}) as to extract face features and train the same classifier on the developed image-identity database by themselves. Without any overlapping with the images used for labeling, the test sets are collected in different days. In total, the test set contains $5,580$ face images collected in the UK and $2,840$ face images collected in China.

\subsubsection{Performance} 
\label{ssub:face_identification}
Fig.~\ref{fig:identification} compares the face identification results of both \sysname and the competing approaches. Face identification is a multi-class classification task and aims to find an unknown person in a dataset of POI. As we can see, by using the face database developed by \sysname, the identification accuracy quickly saturates and is able to have no errors within three guesses (rank-3) on both datasets. For the UK dataset where there are $20$ POI, the rank-1 accuracy of \sysname can be as high as 95.8\% and outperforms the best competing approach (OA) by $12.5\%$. The advantage of \sysname is more significant on the CHN dataset, and it surpasses the best competing approach OA by $\sim 19\%$ (98.0\% vs 79.1\%). In addition, although D-AutoTune's performance is inferior to \sysname, it is still more accurate than competing approaches, especially on the UK dataset. We note that these results are consistent with face labeling results in \sect\ref{ssub:overall_labeling_performance}. Overall, face identification by \sysname are highly accurate, considering that \sysname is only supervised by the weak and noisy device presence information.

\begin{figure*}[t!]
	\centering
	\begin{subfigure}{0.31\textwidth}
	\includegraphics[width=\linewidth]{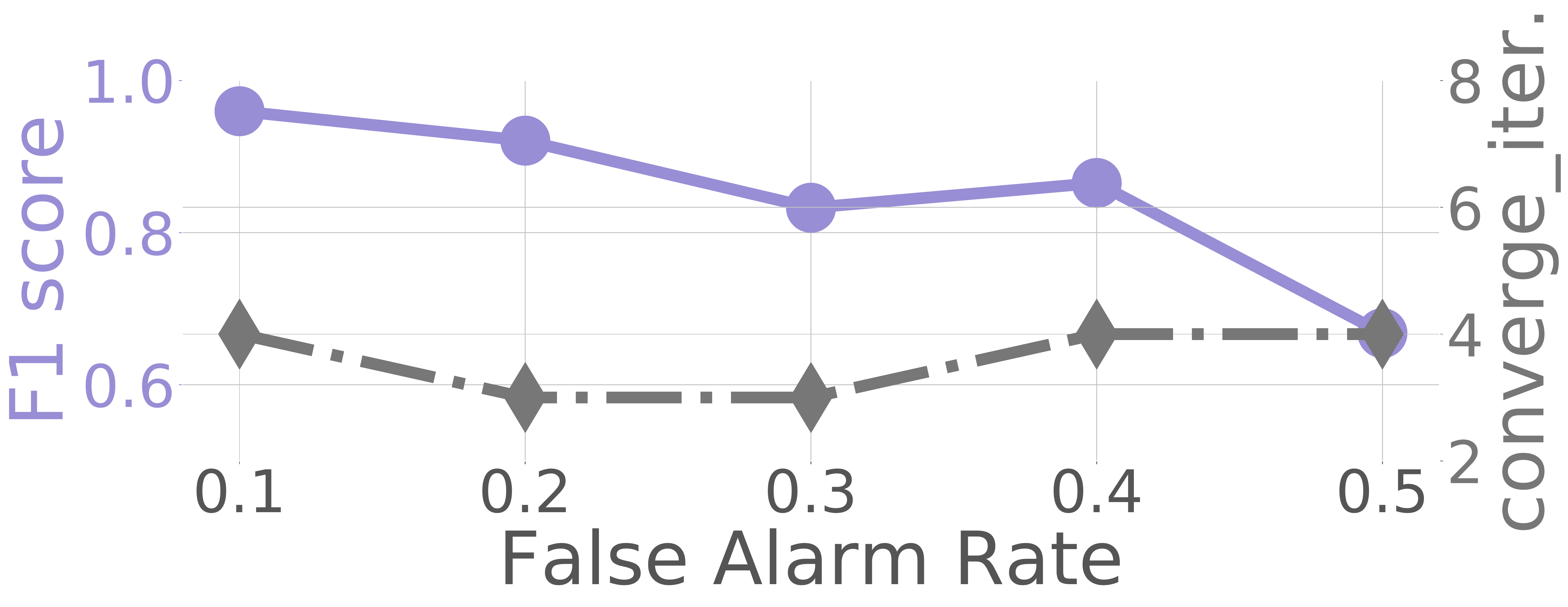}
	\caption{False-alarm Faces} \label{fig:error_1_autotune}
	\end{subfigure}\hspace*{\fill}
	\begin{subfigure}{0.31\textwidth}
	\includegraphics[width=\linewidth]{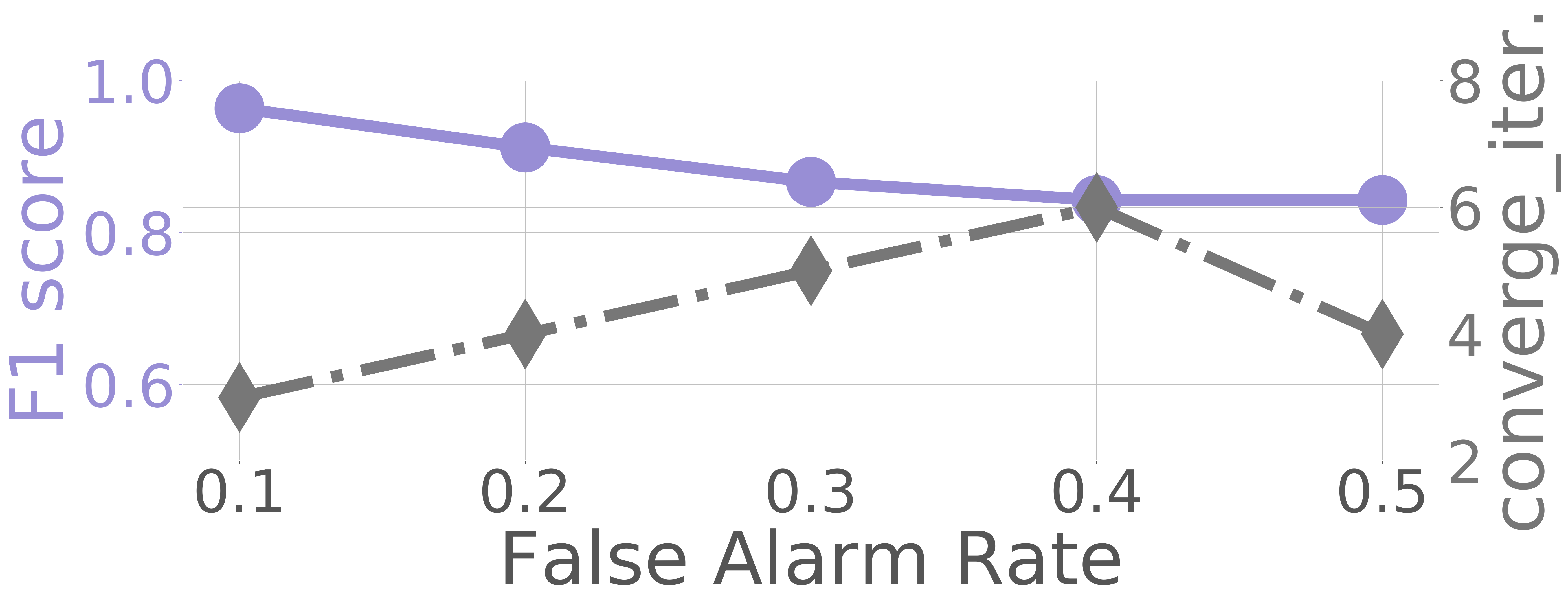}
	\caption{False-alarm Devices} \label{fig:error_2_autotune}
	\end{subfigure}\hspace*{\fill}
	\begin{subfigure}{0.31\textwidth}
	\includegraphics[width=\linewidth]{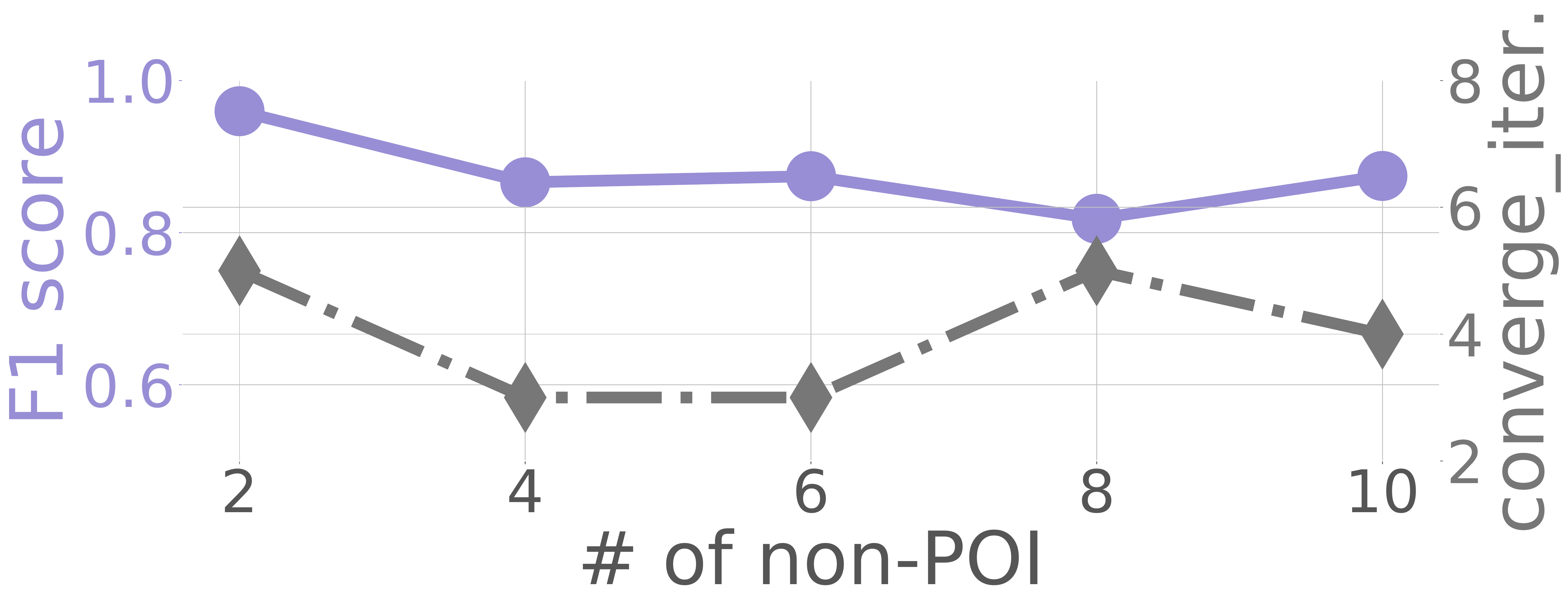}
	\caption{Non-POI Disturbance} \label{fig:error_3_autotune}
	\end{subfigure}

\caption{Results on simulation data showing the impact of sources of noise} \label{fig:result_simulation}
\end{figure*}

\subsection{Sensitivity Analysis} 
\label{sub:robust_analysis_in_simulation_environment}
To further study the performance of \sysname under different noise conditions, we also conduct extensive sensitivity analysis via simulation, considering three types of noise that are common in real-world settings:
\eat{. The benefit of simulation is that we have the full control over the data, and we can alter various settings to validate \sysname under noisy conditions. From the above real-world experimental results, we observed the following three types of noise in an event that impact the performance of \sysname:}

	\noindent \textbf{False-alarm Faces} is the case that the number of detected faces of POI is \emph{greater than} the number of distinct sniffed MAC addresses. 

	\noindent \textbf{False-alarm Devices} is the case that the number of detected faces of POI is \emph{smaller than} the number of distinct sniffed MAC addresses. 

	\noindent \textbf{Non-POI Disturbance} is the case that there are detected faces that belong to a non-POI, whose MAC address are not in our database i.e. they should be discarded as an outlier.
    
\subsubsection{Data Synthesis} 
\label{ssub:data_preparation}
Although a number of public databases of face images are available, unfortunately, none of them come with device presence observations, i.e., sniffed MAC addresses. As an alternative, we develop a synthetic dataset on top of the widely used PIE face database\footnote{http://www.cs.cmu.edu/afs/cs/project/PIE/MultiPie/Multi-Pie/Home.html}. In total $40$ subjects are randomly selected, in which $30$ of them are POI and the rest are mixed in to make the dataset realistic. We then assign to these ``events'' noisy identity observations to simulate device presence information. Based on different types of noise, different synthetic datasets are generated on which \sysname is examined. On average, each subject in our simulation has $168$ images after data augmentation \cite{krizhevsky2012imagenet}. As there is no actual device presence, we simulate this by first randomly placing the face images into different ``events'', in which multiple subjects ``attend''. As the number of images of subjects might be skewed in events, we adopt the $F_1$ score as the metric to evaluate the \sysname's performance.

\subsubsection{False-alarm Faces} 
\label{ssub:false_alarm_faces}
WiFi sniffing is an effective approach to detect the presence of users' mobile devices, however, such detection is not guaranteed to perfectly match the user's presence. For instance, a device could be forgotten at home, be out of battery or simply the WiFi function might be turned off. In the simulation, we vary the error rate of such false alarm faces from $0.1$ to $0.5$. Error rate at $0.1$ means that on average, 10\% of the detected faces in each event are false detections. Fig.~\ref{fig:error_1_autotune} shows the $F_1$ score and convergence iterations of \sysname under different levels of such noises. As we can see \sysname tolerates false-alarm faces well and is able to keep the $F_1$ score above $0.83$ when the false alarm rate is below $0.4$, though it degrades to $0.67$ when the rate rises up to $0.5$. However, we found such case, i.e., on average half of WiFi MAC addresses are missed in all meetings, is rare in the real-world. Finally, we found that false-alarm faces do not affect the convergence and \sysname quickly converges within $4$ iterations in all the cases. 

\subsubsection{False-alarm Devices} 
\label{ssub:false_alarm_devices}
Though surveillance cameras are becoming increasingly ubiquitous, there are cases where the subjects are not captured by the cameras e.g. duto occlusions. This becomes an instance of device false alarm, if her device MAC address is still sniffed. We vary the rate of such false alarm devices from $0.1$ to $0.5$, where $0.1$ means that on average, 10\% of the detected devices in each event are false detections. Fig.~\ref{fig:error_2_autotune} shows that although the $F_1$ score of \sysname decreases, it degrades slowly and stops at $0.84$ after the false-alarm rate becomes $0.5$. As the injected noise becomes stronger, \sysname needs more iterations to converge. However, the largest convergence iteration is still below $7$ (rate at $0.4$). Overall, \sysname is very robust to such type of noises. 

\subsubsection{Non-POI Disturbance} 
\label{ssub:non_poi_disturbance}
Non-POI disturbance happens when subjects without registered MAC addresses  are captured by the camera. We found such noise dominates all the three types of errors. We vary the number of non-POI from $2$ to $10$ and the probability of each non-POI's presence in an event is set to $0.1$. Fig.~\ref{fig:error_3_autotune} shows that \sysname does not suffer much from mild disturbance (2 non-POI), and the $F_1$ score drops slowly to $0.87$ with larger disturbance ($4$ non-POI). In addition in all cases \sysname quickly converges within 5 iterations. 

\section{Discussion and Future Work} 
\label{sec:discussion_and_future_work}

This section discusses some important issues related to AutoTune.

\noindent \textbf{Overheads: } Compared with conventional face recognition methods, AutoTune incurs overheads due to FaceNet fine-tuning. In our experiment, fine-tuning is realized on one NVIDIA K80 GPU. One fine-tuning action takes around $1.2$ hours and $0.5$ hours for the UK dataset and CHN dataset respectively. As we discusses in \sect\ref{sub:offline_cross_modality_face_labeling}, AutoTune is able to converge within $5$ iterations for most of the time, depending on the hyper-parameter setting. Therefore, fine-tuning overheads can be controlled in $7$ hours for the UK dataset and $2.5$ hours for the CHN dataset. Compared to the FaceNet pre-training that takes days and requires more GPUs, the fine-tuning costs of AutoTune is much cheaper. In future work, we will look into an online version of AutoTune, which can incrementally fine-tune the network on the fly when the data is streaming in.

\noindent \textbf{Privacy: } In practice, AutoTune requires face images and device ID of users to operate, which may have certain impacts on user privacy. For example, a user could be identified without explicit consent in a new environment, if the owner has the access to the face image of this user. In this work, we do not explicitly study the attack model in this context, we note that potential privacy concerns are worth exploring in future work.

\section{Conclusion} 
\label{sec:conclusion}
In this work, we described \sysname, a novel pipeline to simultaneously label face images in the wild and adapt a pre-trained deep neural network to recognize the faces of users in new environments. A key insight that motivates it is that enrolment effort of face labelling is unnecessary if a building owner has access to a wireless identifier, e.g., through a smart-phone's MAC address. By learning and refining the noisy and weak association between a user's smart-phone and facial images, \sysname can fine-tune a deep neural network to tailor it to the environment, users, and conditions of a particular camera or set of cameras. Particularly, a novel soft-association technique is proposed to limit the impact of erroneous decisions taken early on in the training process from corrupting the clusters.Extensive experiment results demonstrate the ability of \sysname to design an environment-specific, continually evolving facial recognition system with entirely no user effort even if the face and WiFi observations are very noisy.

\begin{acks}
The authors would like to acknowledge the support of the EPSRC through grant EP/M019918/1 and Oxford-Google DeepMind Graduate Scholarship.
\end{acks}

\bibliographystyle{ACM-Reference-Format}
\balance
\bibliography{reference}


\end{document}